\documentclass[11pt]{article}

\usepackage{sectsty}
\usepackage{graphicx}

\usepackage{times}
\usepackage{natbib}
\usepackage[]{graphicx}
\chardef\bslash=`\\ 

\usepackage{amsmath}
\usepackage{amssymb}
\usepackage{algorithm}
\usepackage{algpseudocode}
\usepackage{hyperref}
\usepackage{authblk}

\newtheorem{assumptions}{Assumption}[section]
\newtheorem{propo}{Proposition}
\newtheorem{theorem}{Theorem}
\newtheorem{lemma}{Lemma}
\newtheorem{proof}{Proof}

\addcontentsline{toc}{section}{Acknowledgement}

\providecommand{\keywords}[1]
{
  \small	
  \textbf{\textit{Keywords: }} #1
}

\title{A network-constrain Weibull AFT model for biomarkers discovery}
\author[1]{Claudia Angelini}
\author[2]{Daniela De Canditiis}
\author[1]{Italia De Feis}
\author[3]{Antonella Iuliano\footnote{Corresponding author: {\sf{antonella.iuliano@unibas.it}}}}

\affil[1]{\footnotesize Istituto per le Applicazioni del Calcolo ``M.Picone'' (CNR), Via Pietro Castellino, 111, 80131 Napoli (Italy)
}

\affil[2]{Istituto per le Applicazioni del Calcolo ``M.Picone'' (CNR), Via dei Taurini, 19, 00185 Roma (Italy) 
}

\affil[3]{Department of Mathematics, Computer Science, and Economics, University of Basilicata, Viale dell'Ateneo Lucano, n. 10, 85100, Potenza (Italy)
}
\date{}

\begin{document}
\maketitle	

\begin{abstract}
We propose AFTNet, a novel network-constraint survival analysis method based on the Weibull accelerated failure time (AFT) model solved by a penalized likelihood approach for variable selection and estimation. When using the log-linear representation, the inference problem becomes a structured sparse regression problem for which we explicitly incorporate the correlation patterns among predictors using a double penalty that promotes both sparsity and grouping effect. Moreover, we establish the theoretical consistency for the AFTNet estimator and present an efficient iterative computational algorithm based on the proximal gradient descent method. Finally,  we evaluate AFTNet performance both on synthetic and real data examples.
\end{abstract}








\section{Introduction}


In the last twenty years, the development of high-throughput technology has produced a large amount of heterogeneous biomolecular data that, together with clinical information about patients,  promises to identify stable and interpretable biomarkers to predict survival and characterize personalized therapy. 
Thereby, there has been a growing interest in developing methods for high-dimensional data that integrate genome-scale knowledge into regression models for survival data to create a comprehensive view of molecular mechanisms and disease progression. In this context, the Cox proportional hazard model \citep{cox1972regression}, which assumes a constant hazard ratio over time, has been naturally extended to deal with the so-called problem $p>>n$, representing $p$ the number of regressors and $n$ the number of samples, see the pioneeristic papers by \citep{fan2002variable, gui2005penalized, tibshirani1997lasso, zhanglu2007, duetal2010, benneretal2010, antoniadisetal2010}. At the same time, the availability of databases encoding information about genes/proteins regulatory mechanisms and pathways demanded more advanced techniques to integrate this knowledge into the models. Network-penalized Cox regression methods explicitly incorporated the relationships among the variables in the penalty term, improving the prediction capabilities and better addressing the inherent structure of omics data \citep{wangetal2009, kimetal2012, wuwang2013, zhang2013network, gongetal2014, huangetal2014, sun2014network, huangetal2016, iuliano2016cancer, verissimo2016degreecox, iuliano2018combining, jiangliang2018, iuliano2021cosmonet, Li et al2021}.

An attractive alternative to the Cox model is the accelerated failure time (AFT) model, where the covariates can accelerate or decelerate the life course of an event by some constant. Different papers extended the AFT model to the high-dimensional case, considering semiparametric estimators based on weighted least squares or rank-based losses. \cite{huangetal2006} and \cite{huangma2010} introduced the regularized Stute's weighted least squares (LS) estimator combining LASSO \citep{tibshirani1996regression}, the threshold-gradient-directed regularization method \citep{friedmanpopescu2004} and the bridge penalty \citep{frankfriedman1993}. \cite{huangharrington2005, dattaetal2007, wangetal2008} and \cite{khanshaw2013} presented the regularized Buckley–James estimator using LASSO, elastic net \citep{zou2005regularization} and Dantzig selector \citep{candes2007dantzig}. \cite{shaetal2006} developed a Bayesian variable selection approach;  \cite{englerli2009} and \cite{caietal2009} developed the regularized Gehan's estimator considering LASSO and the elastic net penalties. Recently, \cite{chengetal2022} considered the Stute's weighted least squares criterion combined with the \textit{l}$_0$-penalty extending the support detection and root finding (SDAR) algorithm \citep{huangetal2018} for linear regression model to the AFT model. 

In the context of network penalized AFT models, \cite{renetal2019} developed a robust network-based variable selection method 
based on the regularized Stute's least absolute deviation (LAD) estimator with a penalty of an $MCP + \textit{l}_1$ form. MCP is the Minimax Concave Penalty \citep{zhang2010} encouraging sparsity, and the $\textit{l}_1$ term promotes the network structure incorporating the network adjacency through the Pearson correlation coefficient.
Recently, \cite{suder2022scalable} proposed a new alternating direction method of multipliers algorithm based on proximal operators for fitting semiparametric AFT models. They minimize a penalized Gehan's estimator considering both the weighted elastic net and the weighted sparse group lasso as penalties. The {\tt R} package  {\tt penAFT} (\url{https://cran.r-project.org/web/packages/penAFT/index.html}) implements the proposed approach.

The literature on high-dimensional parametric AFT models is less extensive than semiparametric ones. However, these models offer an interesting alternative to weighted least squares or rank-based estimators since they are simple, relatively robust against the misspecification of the assumed distribution \citep{huttonmonaghan2002}, and maximum likelihood estimation can be used for inference. \cite{park2018penalized} proposed LASSO, adaptive lasso \citep{zou2006adaptive} and SCAD \citep{fan2001variable} for both the log-normal and the Weibull AFT models. \cite{barnwal2022survival} presented an interesting implementation of the AFT model using XGBoost \citep{xgboost}, a widely used library for gradient boosting, considering the log-normal, log-logistic, and the Weibull AFT models. \cite{alametal2022} penalized the log-likelihood function with Firth's penalty term \citep{firth1993} to overcome the problems due to small sample or rare events for the log-normal, log-logistic and the Weibull AFT models. 

To the best of the authors' knowledge, the problem of incorporating biomolecular knowledge into high-dimensional parametric AFT models still needs to be addressed.  In this paper, we fill this gap with AFTNet, a novel method for the Weibull AFT model based on a double penalty that combines LASSO and quadratic Laplacian penalties \citep{lili2010},  to promote both sparsity and grouping effect. When using the log-linear representation, the inference becomes a structured sparse regression problem for which we implement an efficient iterative computational algorithm based on the proximal gradient descent method and cross-validated linear predictors approach (CV-PL) \citep{dai2019cross}. Moreover, we establish the proposed estimator's theoretical consistency and evaluate its performance on synthetic and real data examples.

The papers is organized as follows. Section \ref{section:MB} presents the mathematical background. Section \ref{section:Inference} introduces the AFTNet estimator and its theoretical property. Section \ref{section:opt} discusses the numerical implementation. Results are shown in Sections \ref{section: simul}-\ref{section: real}, and conclusions are drawn in Section \ref{conclusions}.

 \section{Mathematical background}\label{section:MB}
 
The AFT model is a flexible mathematical framework that describes the relationship between a set of covariates and a time-to-event response. In this model, the non negative random variable $T$ describing the time-to-event is related to covariates  $\boldsymbol{x}=(x_1,\ldots,x_p)^T \in \mathbb{R}^p$ through the hazard function 
 \begin{equation} \label{eq: hazard di T}
 	h(t)= h_0\left(t e^{-\boldsymbol{x}^\top \boldsymbol{\beta}}\right) e^{-\boldsymbol{x}^\top \boldsymbol{\beta}},    
 \end{equation}
 where $\boldsymbol{\beta}=(\beta_1,\ldots,\beta_p)^T$ is the parameter vector, $\boldsymbol{x}^\top \boldsymbol{\beta}$ denotes the joint effect of covariates and $h_0(\cdot)$ is the baseline hazard function. The baseline hazard function represents the hazard without the effects of covariates, i.e., when $\boldsymbol{x} = (0, \ldots, 0)$. 
The  hazard function in Eq.\,\eqref{eq: hazard di T} explains the name AFT model. Whereas in a proportional hazards (PH) model, the covariates act multiplicatively on the hazard, in an AFT model the covariates act multiplicatively on time.  
 Then in AFT models, the effect of covariates is such that if $e^{-\boldsymbol{x}^\top \boldsymbol{\beta}} > 1$, a deceleration of the survival (time) process ensues and if $e^{-\boldsymbol{x}^\top \boldsymbol{\beta}} < 1$, then an acceleration of the survival (time) process occurs. The term $e^{\boldsymbol{x}^\top \boldsymbol{\beta}}$ is known as the accelerated factor.

 Given Eq.\,\eqref{eq: hazard di T}, the corresponding cumulative hazard function $H(t)$ at time $t$ is
 \begin{equation} \label{eq: hazardCum di T}
 	H(t)=\int_{0}^t h(u)du=-\log\left[S\left(t\right)\right], \quad t \geq 0
 \end{equation}
 where 
 \begin{equation} \label{eq: survival di T}
 	S(t)=S_0\left(t e^{-\boldsymbol{x}^\top \boldsymbol{\beta}}\right)
 \end{equation}
 is the survival function and $S_0(\cdot)$ the baseline survival function. Therefore, we can express the density $f(t)$ as follows:
 \begin{equation*}
 	f(t)=- \frac{d S(t)}{dt}=h(t)e^{-H(t)}, \quad t \geq 0,
 \end{equation*} 
 and the hazard function as follows:
 \begin{equation*} \label{eq: h e S per la T}
 	h(t)=- \frac{d \log (S(t))}{dt}, \quad t \geq 0.
 \end{equation*}

Let now consider a study with a number $n$ of individuals from a homogeneous population and suppose that for each individual, the values of $p$ explanatory variables $\boldsymbol{x}_i =(x_{1i}, \ldots,x_{pi})^T$  have been recorded, $i=1,\dots,n$. 
 Let $T_i \geq 0$ be the survival time (failure time) for each $i=1,\ldots, n$.
 Survival times may be subject to censoring, i.e., the endpoints of one or more individuals are not observed. 
 Usually, the censoring time are i.i.d. non-negative random variables $C_i$, $i=1, \dots, n$.
 The censoring mechanism can be right or left. Right censoring happens after the subjects entered the study; left censoring means that the actual survival time is less than observed. The latter case is less common than the first. We assume only the right censoring can occur, and we make the following hypotheses on the nature of the censoring mechanism. 
 \begin{enumerate}
 	\item[(A1)] Given covariates $\boldsymbol{x}_i$, $T_i$'s and $C_i$'s are conditionally independent and the pairs ($T_i, C_i$)'s are also conditionally independent for $i = 1, \ldots, n$;
 	\item[(A2)] Given covariates $\boldsymbol{x}_i$, $C_i$'s are conditionally non-informative about $T_i$'s. 
 \end{enumerate}
  
With the introduction of the censoring variable, the observations can be represented by the pairs $(\xi_i, \delta_i)$, where $\xi_i=\min(T_i,C_i)$ and $\delta_i=0$ if the $i$-th individual is censored, $\delta_i=1$ if not. 

 In this case, each subject $i=1, \ldots, n $, contributes to the probability density function of the observed sample by $f (t_i) =f(\xi_i)$ if it not censored and by $S(c_i)=S(\xi_i)$ if it is censored. Then, the 
 AFT likelihood function is 
 \begin{equation} \label{eq: likelihood alternative}
 	\prod_{i=1}^n [f(\xi_i)]^{\delta_i}[S(\xi_i)]^{1-\delta_i} =\prod_{i=1}^n \left[\frac{f(\xi_i)}{S(\xi_i)} \right]^{\delta_i} S(\xi_i) =\\
 	\prod_{i=1}^n[h(\xi_i)]^{\delta_i}S(\xi_i)
 	=\prod_{i=1}^n[h\left(\xi_i\right)]^{\delta_i} e^{-H(\xi_i)},
 \end{equation}
 where we have used the definition of hazard function $h(\cdot)=\frac{f(\cdot)}{S(\cdot)}$ and the expression $S(\cdot)=e^{-H(\cdot)}$ from Eq.\,\eqref{eq: hazardCum di T}. Although simple,  the likelihood written as in Eq.\,\eqref{eq: likelihood alternative}, is not suited for applications.  The equivalent form obtained by the log-linear model presented in the following section is preferred. 
 
 \subsection{The log-linear AFT model}\label{section:logAFT}
 
AFT models are unified by adopting a log-linear representation of the model as described in the sequel. This representation
 shows that the AFT model for survival data is closely related to the general linear model used in regression analysis. Moreover, most computer software packages adopt the log-linear AFT form when fitting the data.
The log-linear representation of the AFT model describes a linear relationship between the logarithm of survival time and covariates given by:
 \begin{equation}
 	\log T_i =\boldsymbol{x}_i^\top \boldsymbol{\beta}+ \sigma \varepsilon_i, \quad i=1, \ldots, n,
 	\label{eq:logT}
 \end{equation}
 where $\varepsilon_i  \in \mathbb{R}$ is a random error independent from $\boldsymbol{x}_i$, $\boldsymbol{\beta}=(\beta_1, \beta_2, \ldots, \beta_p)^T$ is the parameter vector and $\sigma$ is the scale parameter. 
  Passing on the time scale from Eq.\, \eqref{eq:logT}, we get 
 \begin{equation*}
 	T_i =e^{\boldsymbol{x}_i^\top \boldsymbol{\beta}} e^{\sigma \varepsilon_i}, \quad i=1, \ldots, n.
 	\label{eq: timescaleequation}
 \end{equation*}
 Hence, the survival function of variable $T_i$ is given by
 \begin{equation} \label{eq: survival alternativa}
 	S_i(t)=P(T_i \geq t) = P( e^{\sigma \varepsilon_i} \geq t e^{-\boldsymbol{x}_i^\top \boldsymbol{\beta}}) = S_0 ( t e^{-\boldsymbol{x}_i^\top \boldsymbol{\beta}}),
 \end{equation}
 where $S_0(t)=P( e^{\sigma \varepsilon_i} \geq t )$ is the survival function of an individual for whom $\boldsymbol{x}=0$. Expression \eqref{eq: survival alternativa}  coincides with expression \eqref{eq: survival di T}, which is the general form of the survival function for the $i$th individual in an AFT model. Moreover, taking the logarithm of both side of Eq.\,\eqref{eq: survival alternativa}, multiplying by $-1$, and differentiating with respect to $t$, we get 
 \begin{equation*}
 	h_i(t)= - \frac{d \log S_i(t)}{dt} = h_0(t e^{-\boldsymbol{x}_i^\top \boldsymbol{\beta}})e^{- \boldsymbol{x}_i^\top \boldsymbol{\beta}},
 \end{equation*}
 which is the general form of the hazard function for the  $i$th individual in an AFT model, see Eq.\,\eqref{eq: hazard di T}. For completeness, in the following, we give the survival, density and hazard functions of the $i$th individual under the AFT model in terms of the random variable $\varepsilon $. These alternative formulations will shortly be used to rewrite the likelihood. 
 The  survival function of the $i$th individual is 
 \begin{equation} \label{eq: S}
 	S_i(t)=P(T_i \geq t)=P(\log T_i \geq \log t)= P \left( \varepsilon_i \geq \frac{\log t -\boldsymbol{x}_i^\top \boldsymbol{\beta} }{\sigma} \right) =S_{\varepsilon_i} \left(\frac{\log t -\boldsymbol{x}_i^\top \boldsymbol{\beta} }{\sigma} \right),
 \end{equation}
 the density function of the $i$th individual is
 \begin{equation} \label{eq: f}
 	f_i(t)= - \frac{d S_i(t)}{d t}= \frac{1}{t \sigma} f_{\varepsilon_i} \left(\frac{\log t -\boldsymbol{x}_i^\top \boldsymbol{\beta} }{\sigma} \right),
 \end{equation}
 the hazard function of the $i$th individual is
 \begin{equation} \label{eq: h}
 	h_i(t)=\frac{1}{t \sigma} h_{\varepsilon_i } \left(\frac{\log t -\boldsymbol{x}_i^\top \boldsymbol{\beta} }{\sigma} \right).
 \end{equation}
 %
 %
 %
 Let denote $y_i=\min(\log T_i,\log C_i)$.  Then,  under assumptions (A1) and (A2),  from  the expressions in Eqs.\ (\ref{eq: S}-\ref{eq: h}), we can express the AFT likelihood function as:
 \begin{eqnarray*}
 	\begin{aligned}
 		&\prod_{i=1}^n \left[\frac{1}{t_i \sigma} f_{\varepsilon_i }\left( \frac{y_i-\boldsymbol{x}_i^\top \boldsymbol{\beta}}{\sigma}\right) \right]^{\delta_i}\left[S_{\varepsilon_i } \left( \frac{y_i-\boldsymbol{x}_i^\top \boldsymbol{\beta}}{\sigma}\right) \right]^{1-\delta_i} \\
 		&= \prod_{i=1}^n \left[\frac{1}{t_i \sigma}  h_{\varepsilon_i } \left( \frac{y_i-\boldsymbol{x}_i^\top \boldsymbol{\beta}}{\sigma}\right) \right]^{\delta_i} S_{\varepsilon_i } \left( \frac{y_i-\boldsymbol{x}_i^\top \boldsymbol{\beta}}{\sigma}\right) ,    
 	\end{aligned}
 	\label{eq:prod}
 \end{eqnarray*}
 and  the AFT log-likelihood (a part the term $\sum_{i=1}^n \delta_i \log t_i$) by 
 \begin{equation} \label{eq:loglike_general}
 	l(\boldsymbol{\beta}^\top, \sigma)=  \sum_{i=1}^n \delta_i \left( \log(f_{\varepsilon_i }(e_i) -\log(\sigma) \right) + (1-\delta_i) \log (S_{\varepsilon_i }(e_i)),
 \end{equation}
 with $e_i= \frac{y_i-\boldsymbol{x}_i^\top \boldsymbol{\beta}}{\sigma}$ the standardized residuals. 
 
 Both  $\boldsymbol{\beta}$ and $\sigma$ are the unknown parameters we aim to estimate.  Let us denote $\boldsymbol{\theta}=(\boldsymbol{\beta}^\top, \sigma)^\top$. 
 Then, the  gradient vector of $l(\boldsymbol{\theta})=l(\boldsymbol{\beta}^\top, \sigma)$ has the following expression:
 \begin{equation}  \label{eq: grad}
 	\left\{ \begin{array}{cl} 
 		\frac{\partial l(\boldsymbol{\theta})}{\partial \beta_j} =& \frac{1}{\sigma} \sum_{i=1}^n a_i x_{ij},   \quad \quad \mbox{for} ~ j=1, \ldots, p,\\
 		 & \\
 		 \frac{\partial l(\boldsymbol{\theta})}{\partial \sigma} = &\frac{1}{\sigma} \sum_{i=1}^n (e_i a_i -\delta_i),
 	\end{array}
 	\right.   
 \end{equation}
 \vspace{0.5 cm}
 with 
 \begin{equation*} \label{eq: ai}
 	\begin{array}{ccl}
 		a_i &=&- \delta_i \frac{d \log (f_{\varepsilon_i }(e_i))}{d e_i}  -(1-\delta_i) \frac{d \log(S_{\varepsilon_i }(e_i))}{d e_i} \\
 		& & \\
 		& =&-\delta_i \frac{d \log (f_{\varepsilon_i }(e_i))}{d e_i} +(1-\delta_i) h_{\varepsilon_i }(e_i).\\
 	\end{array}
 \end{equation*}
 %
 The observed information matrix $I(\boldsymbol{\theta})=I(\boldsymbol{\beta},\sigma)$ has entries
 \begin{equation*}\label{eq:hess}
 	\begin{aligned}
 		&-\frac{\partial l^2(\boldsymbol{\theta})}{\partial \beta_j \partial \beta_k} = \frac{1}{\sigma^2} \sum_{i=1}^n x_{ij} x_{ik} A_i,\\
 		&-\frac{\partial l^2(\boldsymbol{\theta})}{\partial \beta_j \partial \sigma} = \frac{1}{\sigma^2} \sum_{i=1}^n x_{ij} e_{i} A_i + \frac{1}{\sigma} \frac{\partial l(\boldsymbol{\theta})}{\partial \beta_j},\\
 		&-\frac{\partial l^2(\boldsymbol{\theta})}{\partial^2 \sigma} = \frac{1}{\sigma^2} \sum_{i=1}^n (e_i^2 A_i + \delta_i) + \frac{2}{\sigma}   \frac{\partial l(\boldsymbol{\theta})}{\partial \sigma},
 	\end{aligned}
 \end{equation*}
 with $i,k=1,\ldots,p$, and  $A_i=\frac{d a_i}{d e_i} = \delta_i \frac{d^2 \log(f_{\varepsilon_i }(e_i))}{d e_i^2} + (1-\delta_i) [  h_{\varepsilon_i }(e_i) \frac{d \log(f_{\varepsilon_i }(e_i))}{de_i} + h_{\varepsilon_i }^2(e_i)]$. 
 
This is a very general framework.  We obtain different models depending on the distribution specified for $\varepsilon_i$ in Eqs.\ (\ref{eq: S}-\ref{eq: h}). The members of the AFT model class include the exponential AFT model, Weibull AFT model, log-logistic AFT model, log-normal AFT model, and gamma AFT model. Here, we consider the Weibull AFT model presented in the following subsection.
 
 \subsection{The Weibull AFT model}\label{section:Weib}
 
The Weibull model is obtained when the variables $\varepsilon_i$ in Eq.\,(\ref{eq:logT})  are i.i.d.  as a standard Gumbel variable (Extreme Value variable) with density, survival, and hazard functions given by 
 \begin{equation} \label{eq: gumbel distr}
 	f_{\varepsilon_i}(x)=\exp{(x-e^{x})}; \quad S_{\varepsilon_i}(x)= \exp{(-e^{x})};  \quad  h_{\varepsilon_i}(x)=e^{x}; \quad \quad x \in \mathbb{R}.
 \end{equation}
 In this case, it is well known that the variable $ T_i=\exp(\boldsymbol{x}_i^\top \boldsymbol{\beta}+ \sigma \varepsilon_i)$, obtained from the log-linear model of Eq.\,(\ref{eq:logT}), is a Weibull distribution with scale parameter  $\exp(\boldsymbol{x}^\top \boldsymbol{\beta})$ and shape parameter $\frac{1}{\sigma}$, (i.e., $T\sim W\large(\exp(\boldsymbol{x}^\top \boldsymbol{\beta}\large), \frac{1}{\sigma} )$), the proof is given for example in \cite{liu2018using}. The Weibull distribution is the unique distribution (along with its special cases, like the exponential distribution) that satisfies both the Proportional Hazard (PH) and AFT assumptions.
 Moreover, in this case, we can specialize the general model given in the previous section since the standardized residual  $e_i= \frac{y_i-\boldsymbol{x}_i^\top \boldsymbol{\beta}}{\sigma}$ has density, survival and hazard functions given in Eq.\,(\ref{eq: gumbel distr}). Hence, substituting  $\log(f_{\varepsilon_i } (e_i))=e_i-e^{e_i}$ and $\log(S_{\varepsilon_i } (e_i))=-e^{e_i}$ into Eq.\,(\ref{eq:loglike_general}), we get the log-likelihood function for the Weibull AFT model
 \begin{equation} \label{eq: loglike per weibull}
 	\ell(\boldsymbol{\theta})= \sum_{i=1}^n \delta_i \left( - \log (\sigma) + \frac{y_i-\boldsymbol{x}_i^\top \boldsymbol{\beta}}{\sigma} \right) - \exp \left(\frac{y_i-\boldsymbol{x}_i^\top \boldsymbol{\beta}}{\sigma} \right)
 \end{equation}
 as well as, we get the expression of the gradient given in Eq.\,(\ref{eq: grad})  with parameter $a_i$ given by
 \begin{equation} \label{eq: ai per weibull}
 	a_i=-\left[  \delta_i -\exp \left(\frac{y_i-\boldsymbol{x}_i^\top \boldsymbol{\beta}}{\sigma} \right)  \right].   
 \end{equation}
 More precisely, in the case of Weibull AFT model the log-likelihood gradient vector is given by
 \begin{equation}\label{eq:part}
 	\left\{ \begin{array}{cl} 
 		\frac{\partial \ell(\boldsymbol{\theta})}{\partial \beta_j} =& \frac{1}{\sigma} \sum_{i=1}^n \exp \left(\frac{y_i-\boldsymbol{x}_i^\top \boldsymbol{\beta}}{\sigma} \right) x_{ij} - \delta_i x_{ij},   \quad \quad \mbox{for} ~ j=1, \ldots, p,\\
 		& \\
 		\frac{\partial \ell(\boldsymbol{\theta})}{\partial \sigma} = &\frac{1}{\sigma} \sum_{i=1}^n \delta_i \left(-1-\frac{y_i-\boldsymbol{x}_i^\top \boldsymbol{\beta}}{\sigma} \right) + \exp \left(\frac{y_i-\boldsymbol{x}_i^\top \boldsymbol{\beta}}{\sigma} \right) \frac{y_i-\boldsymbol{x}_i^\top \boldsymbol{\beta}}{\sigma}.
 	\end{array}
 	\right.
 \end{equation}
 %
 %
 %
 Moreover in the case of Weibull AFT model, the observed information matrix, $I(\boldsymbol{\theta})=I(\boldsymbol{\beta},\sigma)$, has entries given by 
 \begin{equation}\label{eq:part1}
 	\begin{aligned}
 		&-\frac{\partial \ell^2(\boldsymbol{\theta})}{\partial \beta_j \partial \beta_k} = \frac{1}{\sigma^2} \sum_{i=1}^n x_{ij} x_{ik} \exp(e_i),\\
 		&-\frac{\partial \ell^2(\boldsymbol{\theta})}{\partial \beta_j \partial \sigma} = \frac{1}{\sigma^2} \sum_{i=1}^n x_{ij} e_{i} \exp(e_i) + \frac{1}{\sigma} \frac{\partial \ell(\boldsymbol{\theta})}{\partial \beta_j},\\
 		&-\frac{\partial \ell^2(\boldsymbol{\theta})}{\partial^2 \sigma} = \frac{1}{\sigma^2} \sum_{i=1}^n \left(e_i^2 \exp(e_i) + \delta_i\right) + \frac{2}{\sigma}  \frac{\partial \ell(\boldsymbol{\theta})}{\partial \sigma}.
 	\end{aligned}
 \end{equation}
 %
 %
 %
 

\section{Inference} \label{section:Inference}

Given a sample of $n$ data satisfying the Weibull AFT model, we propose to estimate parameters $\boldsymbol{\theta}=(\boldsymbol{\beta}^\top, \sigma)^\top \in \mathbb{R}^p \times \mathbb{R}^+$ using a penalized maximum likelihood estimation (MLE). The penalty we impose on the coefficients has two motivations, which are very important in modern data analysis, especially for biomarker discovery. The first is the high-dimension of the problems encountered in applications where often the number of data/individuals (sample size $n$) is smaller than the number of predictors/genes ($p >>n$), so a MLE estimator is unfeasible without constraints. The second motivation is that in some applications (especially in genomic cancer applications), one has prior network-constrained information that is very important to exploit during the inference process. 

Let $(Y, \delta)$, with $Y=\min(\log T, \log C)$  be the observable random variables from the censored Weibull AFT model. 
Given a data sample of size $n$ $(y_i,\delta_i,\boldsymbol{x}_i)$, $i = 1,\ldots,n$, under Assumptions  (A1) and (A2),   Eq.\, (\ref{eq: loglike per weibull}) represent the the log-likelihood function, with $a_i$ given in Eq.\,(\ref{eq: ai per weibull}). 

In these settings,  we propose AFTNet as the following penalized estimator for parameter $\boldsymbol{\theta}$, i.e.,
\begin{equation}
	\widehat{\boldsymbol{\theta}}= \underset{\boldsymbol{\theta}}{\operatorname{argmin}} ~    -\ell(\boldsymbol{\theta})+n\, {\cal P}_{\lambda, \alpha}(\boldsymbol{\beta}),
	\label{eq: penalized MLE definition}
\end{equation}
where 
\begin{equation}
	{\cal P}_{\lambda, \alpha}(\boldsymbol{\beta})=\lambda[\alpha \vert|\boldsymbol{\beta}\vert|_1+(1-\alpha)\Omega(\boldsymbol{\beta})],
	\label{eq: penalty}
\end{equation}
with $\lambda>0$ the regularization parameter and $\alpha \in [0,1]$ a fixed parameter, balancing the convex combination of the two  penalty terms. In the penalty in Eq. (\ref{eq: penalty}), the first term is the classical $\ell_1$-norm, i.e., LASSO penalty, which forces the individual parameter estimate sparsity; the second term is a Laplacian matrix constraint $\Omega(\boldsymbol{\beta})$ that gives smoothness among connected variables in the network encoding prior knowledge. This prior knowledge is encoded by an undirected graph with $p$ nodes associated with the parameter $\beta_j$, $j=1, \ldots,p$, so that the Laplacian  matrix constraint is $\Omega(\boldsymbol{\beta})=\boldsymbol{\beta}^T \boldsymbol{L} \boldsymbol{\beta}$, with $\boldsymbol{L} = \boldsymbol{D} - \boldsymbol{A} \in \mathbb{R}^{p\times p}$  the graph Laplacian, being $\boldsymbol{D}$ the degree matrix and $\boldsymbol{A}$ the adjacency matrix. The Laplacian matrix constraint,  $\Omega(\boldsymbol{\beta})$,  nearly sets all of the connected coefficients in the network to zero or non-zero values. 

We stress that in Eq.\,(\ref{eq: penalty}),  we assume that the Laplacian matrix $\boldsymbol{L}$ is known and available from the literature, the regularization parameter $\lambda>0$ can be estimated using some data-driven model selection procedure, the parameter $\alpha \in [0,1]$ is fixed and it is chosen by the user to balance the two penalties. 

We provide theoretical properties of the  AFTNet estimator in Section \ref{section:thorProp}.  Numerical solution to problem in Eq.\,(\ref{eq: penalized MLE definition}) is demanded to Section \ref{section:opt}


\subsection{Theoretical properties}\label{section:thorProp}

We can not write a closed form expression for the solution to problem \eqref{eq: penalized MLE definition} using the Karush–Kuhn–Tucker (KKT) conditions since the log-likelihood $\ell(\boldsymbol{\theta})$ is convex for variable $ \boldsymbol{\beta}$ and not for variable $\sigma$. For this reason, it is not possible to extend asymptotic results for Cox given in \cite{sun2014network}. Hence, we resort to the finite sample approach proposed in \cite{Loh&Wain2015}, which establishes error bounds when both the loss and penalty are allowed to be non-convex, provided that the loss function satisfies a form of restricted strong convexity and the penalty satisfies suitable mild conditions.


Let us consider the likelihood in Eq.\,(\ref{eq: penalized MLE definition}), $\ell(\boldsymbol{\theta})=\sum_{i=1}^{n} \ell_i(\boldsymbol{\theta})$ with
	\begin{equation*}
		\ell_i(\boldsymbol{\theta})=\delta_i\Bigg(-\log(\sigma)+\frac{y_i-\boldsymbol{x}_i^T \boldsymbol{\beta}}{\sigma}\Bigg)-\exp\Bigg({\frac{y_i-\boldsymbol{x}_i^T \boldsymbol{\beta}}{\sigma}}\Bigg);
	\end{equation*}
	and let us define  $\ell^{(n)}(\boldsymbol{\theta})=-\frac{1}{n}\ell(\boldsymbol{\theta})$.
We have the following gradient vector and Hessian matrix
	\begin{itemize}
	\item $\nabla \ell(\boldsymbol{\theta})=\sum_{i=1}^{n} \nabla  \ell_i(\boldsymbol{\theta}_i)$ where $ \nabla\ell_i(\boldsymbol{\theta}) \in \mathbb{R}^{p+1}$, by  Eq.\,(\ref{eq:part}), is
	\begin{equation} \label{eq:gradiente}
		\nabla \ell_i(\boldsymbol{\theta_j})=\begin{cases}
			\frac{1}{\sigma}\, \exp\Bigg({\frac{y_i-\boldsymbol{x}_i^T \boldsymbol{\beta}}{\sigma}}\Bigg) x_{ij}-\delta_i\, x_{ij}, \quad j=1, \ldots, p,\\
			\frac{1}{\sigma}\, \delta_i \left(-1-\frac{y_i-\boldsymbol{x}_i^T \boldsymbol{\beta}}{\sigma} \right)+\exp\Bigg({\frac{y_i-\boldsymbol{x}_i^T \boldsymbol{\beta}}{\sigma}}\Bigg)\frac{y_i-\boldsymbol{x}_i^T \boldsymbol{\beta}}{\sigma}, 
		\end{cases}
	\end{equation}
	\item $\nabla ^2 \ell(\boldsymbol{\theta})=\sum_{i=1}^{n} \nabla^2  \ell_i(\boldsymbol{\theta})$ where $\nabla^2  \ell_i(\boldsymbol{\theta})\in \mathbb{R}^{p+1} \times \mathbb{R}^{p+1} $, by Eq.\,(\ref{eq:part1}) , is
	\begin{equation*}  \label{eq:hessiano}
		\bigg[\nabla^2 \ell_i(\boldsymbol{\theta})\bigg]_{jk}=-\begin{cases}
			\frac{1}{\sigma^2} x_{ij} x_{ik} \exp(e_i), \quad j,k =1, \ldots, p,\\
			\frac{1}{\sigma^2} x_{ij} e_i \exp(e_i)+\frac{1}{\sigma}\frac{\partial\ell_i}{\partial \beta_j}, \quad j=1, \ldots, p; ~  k=p+1,\\
			\frac{1}{\sigma^2} (e^2_i \exp(e_i)+\delta_i)+\frac{2}{\sigma}\frac{\partial\ell_i}{\partial \sigma}, \quad j=k=p+1.
		\end{cases}
	\end{equation*}
\end{itemize}

Inspired by the work of \cite{InspirationTheoretical}, we make the following assumptions:

\begin{assumptions}\label{assump1}
	\emph{(Bounded data)}.
	\begin{itemize}
		\item[(i)] $\exists \tau_y$: $\lvert Y_i\rvert \leq \tau_y$, for all $i=1, \ldots, n$,
		\item [(ii)]  $\exists \tau_x$: $\lVert x_{ij}\rVert < \tau_x$ $\forall i,j$.
	\end{itemize}
\end{assumptions}
\begin{assumptions}\label{assump2}
	\emph{(Bounded true parameter)}. Let $\boldsymbol{\theta}^*=({\boldsymbol{\beta}^*}^T, \sigma^*)^T \in \mathbb{R}^p \times \mathbb{R}^+$ be the true parameter vector, i.e.,
		\begin{equation*}
		\boldsymbol{\theta}^*=\underset{\boldsymbol{\theta}}{\operatorname{argmin}} ~ \mathbb{E}_{XY} [\ell^{(n)}(\boldsymbol{\theta})],\quad 
		\mathbb{E}_{XY}[ \nabla\ell^{(n)}(\boldsymbol{\theta}^*)]=0,\quad 
		\mathbb{E}_{XY}[ \nabla^2\ell^{(n)}(\boldsymbol{\theta}^*)] \succ 0,
	\end{equation*} 
then $\exists \, R \, \mbox{and} \, s << p$, such that
	\begin{equation}
		\lVert\boldsymbol{\theta}^*\rVert_1 \leq R;  \quad 	\lVert\boldsymbol{\beta}^*\rVert_0=s=\big|\{j:\, \beta_j^* \neq 0\}\big|; \quad \sigma^* >0.
		\label{eq:boundedpar}
	\end{equation}

\end{assumptions}
\begin{assumptions}\label{assump3}
	\emph{(Bounded minimum eigenvalue population Hessian)}:
	\begin{equation*}
		\exists \gamma >0: \min_{\boldsymbol{\theta}:\lVert \boldsymbol{\theta}-\boldsymbol{\theta}^*\rVert_2 \leq 2R}\lambda_{min}  \left( \mathbb{E}[ \nabla^2\ell^{(n)}(\boldsymbol{\theta})]  \right) \geq \gamma,
		\label{eq:boounded hessian}
	\end{equation*}
	with $\lambda_{min} (A)$ being the minimum eigenvalue of matrix $A$.
\end{assumptions}
\vspace{0.5cm}

Suppose matrix $\boldsymbol{L} \succeq 0$. Then,  the solution $\hat{\boldsymbol{\theta}}$ of the minimization problem in 
Eq.\ (\ref{eq: penalized MLE definition}) and Eq.\ (\ref{eq: penalty}) satisfies
\begin{equation*}
	\begin{aligned}
		\hat{\boldsymbol{\theta}}&=\underset{\boldsymbol{\theta}\in \mathbb{R}^p \times \mathbb{R}^+, \lVert \boldsymbol{\theta}\rVert_1 \leq R}{\operatorname{argmin}} ~  -\ell(\boldsymbol{\theta})+n \lambda \alpha \lVert \boldsymbol{\beta}\rVert_1+n \lambda(1-\alpha)\lVert \boldsymbol{L}^{\frac{1}{2}}\boldsymbol{\beta}\rVert^2_2\\
		&=\underset{\boldsymbol{\theta \in \mathbb{R}^p \times \mathbb{R}^+, \, \lVert \boldsymbol{\theta}\rVert_1 \leq R}}{\operatorname{argmin}} ~  -\frac{1}{n}\ell(\boldsymbol{\theta})+\lambda \alpha \lVert \boldsymbol{\beta}\rVert_1+ \lambda(1-\alpha)\lVert \boldsymbol{L}^{\frac{1}{2}}\boldsymbol{\beta}\rVert^2_2\\
		&=\underset{\boldsymbol{\theta \in \mathbb{R}^p \times \mathbb{R}^+, \, \lVert \boldsymbol{\theta}\rVert_1 \leq R}}{\operatorname{argmin}} ~ \ell^{(n)}(\boldsymbol{\theta})+\lambda \alpha \lVert \boldsymbol{\beta}\rVert_1+ \lambda(1-\alpha)\lVert \boldsymbol{L}^{\frac{1}{2}}\boldsymbol{\beta}\rVert^2_2,
	\end{aligned}
	\label{eq:estimationTheta}
\end{equation*}
%
%
Here, we  include the side condition $ \lVert \boldsymbol{\theta}\rVert_1 \leq R$ to guarantee the existence of at least one local/global optima. 

In the following, we state the main result. 

\begin{theorem}\label{theorem1}
	Under Assumptions \ref{assump1},\ref{assump2} and \ref{assump3} 
	for all $\hat{\boldsymbol{\theta}}$ local minimizers, i.e. that satisfy the first order condition
	\begin{equation*}
		\langle \nabla \ell^{(n)}(\hat{\boldsymbol{\theta}})+\nabla {\cal P}_{\lambda, \alpha}(\hat{\boldsymbol{\theta}}), \boldsymbol{\theta}-\hat{\boldsymbol{\theta}}\rangle \geq 0, \quad \forall \boldsymbol{\theta}: \ \lVert \boldsymbol{\theta}\rVert_1 \leq R,
		\label{eq:firstorder}
	\end{equation*}
	it holds 
	\begin{equation*}
		\lVert \hat{\boldsymbol{\theta}}- \boldsymbol{\theta}^*\rVert_2 \leq \frac{\frac{3}{2}\lambda \alpha \sqrt{s+1}+2 \lambda(1-\alpha)R \lambda_{max}(\boldsymbol{L}) }{\gamma}
		\label{eq:firstorder1}
	\end{equation*}
	with high probability.
\end{theorem}
To prove Theorem \ref{theorem1}, we need the following Lemmas whose proofs are given in the Appendix.

\begin{lemma}\label{lemma1}
	(Restricted strong convexity (RSC)) Under Assumptions \ref{assump1},\ref{assump2} and \ref{assump3} it holds: $\exists \gamma>0$ and $\exists  \tau >0$ such that
	\begin{equation*}
		\langle \nabla \ell^{(n)}(\boldsymbol{\theta}^*+\Delta \boldsymbol{\theta})-\nabla\ell^{(n)}(\boldsymbol{\theta}^*),\Delta \boldsymbol{\theta}\rangle \geq \gamma \lVert \Delta \boldsymbol{\theta}\rVert^2_2- \tau\, \sqrt{\frac{\log n(p+1)}{n}} \, \lVert \Delta \boldsymbol{\theta}\rVert_1^2, \quad \mbox{for all} \, \Delta \boldsymbol{\theta}: \,  \lVert \Delta \boldsymbol{\theta}\rVert_1\leq 2R,
		\label{lem1}
	\end{equation*}
	with high probability.
\end{lemma}
\begin{lemma}	\label{lemma2}
	(Bounded gradient) Under Assumptions \ref{assump1}, \ref{assump2}  and \ref{assump3} it holds
	\begin{equation}
		\max \Bigg\{ \lVert \nabla \ell^{(n)}(\boldsymbol{\theta}^*)\rVert_{\infty}, \tau\, R \sqrt{\frac{\log n(p+1)}{n}}\Bigg\} \leq \frac{\lambda \alpha}{4},
		\label{lem2}
	\end{equation}
	with high probability and suitable choice of $\lambda>0$.
\end{lemma}
Now we are able to demonstrate Theorem \ref{theorem1}.

\begin{proof}
Define the error vector $\boldsymbol{\nu}=\hat{\boldsymbol{\theta}}-\boldsymbol{\theta}^*$.  Under Assumption \ref{assump2}, we have $\lVert \boldsymbol{\nu} \rVert_1 \leq \lVert \hat{\boldsymbol{\theta}} \rVert_1 + \lVert \boldsymbol{\theta}^* \rVert_1\leq 2R$, hence by Lemma \ref{lemma1}, it holds the RSC 
	\begin{equation*}
		\langle \nabla \ell^{(n)}(\hat{\boldsymbol{\theta}})-\nabla\ell^{(n)}(\boldsymbol{\theta}^*),\boldsymbol{\nu} \rangle \geq \gamma \lVert \boldsymbol{\nu} \rVert^2_2 - \tau\, \sqrt{\frac{\log n(p+1)}{n}} \, \lVert\boldsymbol{\nu} \rVert_1^2, 
		\label{eq:proof1}
	\end{equation*}
	from which follows that
	\begin{equation}
		 \gamma \lVert \boldsymbol{\nu} \rVert^2_2\leq \langle \nabla \ell^{(n)}(\hat{\boldsymbol{\theta}})-\nabla\ell^{(n)}(\hat{\boldsymbol{\theta}}^*),\boldsymbol{\nu} \rangle +\tau\, \sqrt{\frac{\log n(p+1)}{n}} \, \lVert \boldsymbol{\nu} \rVert_1^2.
		\label{eq:proof2}
	\end{equation}
	Since $ \| \boldsymbol{\theta}^* \|_1 \leq R $, by using the first order condition for  $\hat{\boldsymbol{\theta}}$ we have
	\begin{equation*}
		\langle \nabla \ell^{(n)}(\hat{\boldsymbol{\theta}})+ \nabla {\cal P}_{\lambda, \alpha}(\hat{\boldsymbol{\theta}}), \boldsymbol{\theta}^*-\hat{\boldsymbol{\theta}}\rangle \geq 0
		\label{eq:proof3}
	\end{equation*}
	(note that the penalty is only on $\boldsymbol{\beta}$, so we have implicitly assume that the last element of ${\cal P}_{\lambda}(\hat{\boldsymbol{\theta}})$ is zero), i.e., 
	\begin{equation*}
		\langle \nabla \ell^{(n)}(\hat{\boldsymbol{\theta}})+ \nabla {\cal P}_{\lambda, \alpha}(\hat{\boldsymbol{\theta}}), -\boldsymbol{\nu}\rangle \geq 0,  \quad  \Rightarrow  \quad \langle \nabla \ell^{(n)}(\hat{\boldsymbol{\theta}}),\boldsymbol{\nu}\rangle \leq \langle \nabla {\cal P}_{\lambda, \alpha}(\hat{\boldsymbol{\theta}}), -\boldsymbol{\nu} \rangle,
		\label{eq:proof4}
	\end{equation*}

	and using this inequality in Eq.\,(\ref{eq:proof2}), we have
	\begin{equation*}
		\begin{aligned}
			\gamma \lVert \boldsymbol{\nu} \rVert_2^2 &\leq
			\langle \nabla \ell^{(n)}(\hat{\boldsymbol{\theta}}),\boldsymbol{\nu} \rangle -  \langle\nabla \ell^{(n)}(\boldsymbol{\theta}^*),\boldsymbol{\nu} \rangle  +\tau\, \sqrt{\frac{\log n(p+1)}{n}} \, \lVert \boldsymbol{\nu} \rVert_1^2\\
			& \leq \langle \nabla {\cal P}_{\lambda, \alpha}(\hat{\boldsymbol{\theta}}), -\boldsymbol{\nu} \rangle-  \langle \nabla \ell^{(n)}(\boldsymbol{\theta}^*),\boldsymbol{\nu} \rangle +\tau\, \sqrt{\frac{\log n(p+1)}{n}} \, \lVert \boldsymbol{\nu} \rVert_1^2\\
			&=\langle \lambda \alpha  \nabla \lVert  \hat{\boldsymbol{\beta}} \rVert_1  +\lambda(1-\alpha) \nabla \rVert \boldsymbol{L}^{\frac{1}{2}} \hat{\boldsymbol{\beta}}\rVert_2^2, \boldsymbol{\beta}^*-\hat{\boldsymbol{\beta}}\rangle - \langle\nabla \ell^{(n)}(\boldsymbol{\theta}^*),\boldsymbol{\nu} \rangle+\tau\, \sqrt{\frac{\log n(p+1)}{n}} \, \lVert \boldsymbol{\nu} \rVert_1^2\\
			&=\langle \lambda \alpha \nabla   \lVert \hat{\boldsymbol{\beta}} \rVert_1,\boldsymbol{\beta}^*-\hat{\boldsymbol{\beta}}\rangle+   \langle 2 \lambda(1-\alpha)\boldsymbol{L} \hat{\boldsymbol{\beta}}, \boldsymbol{\beta}^*-\hat{\boldsymbol{\beta}}\rangle- \langle\nabla \ell^{(n)}(\boldsymbol{\theta}^*),\boldsymbol{\nu} \rangle + \tau\, \sqrt{\frac{\log n(p+1)}{n}} \, \lVert \boldsymbol{\nu} \rVert_1^2\\
			& \leq  \langle  \lambda \alpha \nabla  \lVert \hat{\boldsymbol{\beta}} \rVert_1,\boldsymbol{\beta}^*-\hat{\boldsymbol{\beta}}\rangle+   \lvert \langle 2 \lambda(1-\alpha)\boldsymbol{L} \hat{\boldsymbol{\beta}}, \boldsymbol{\beta}^*-\hat{\boldsymbol{\beta}}\rangle\rvert - \langle\nabla \ell^{(n)}(\boldsymbol{\theta}^*),\boldsymbol{\nu} \rangle + \tau\, \sqrt{\frac{\log n(p+1)}{n}} \, \lVert \boldsymbol{\nu} \rVert_1^2.
		\end{aligned}
		\label{eq:proof6}
	\end{equation*}
	Now, since the lasso penalty is convex, using Cauchy-Schwarz inequality and H\"older inequality, we have

	\begin{equation*}
		\begin{aligned}
			\gamma \lVert \boldsymbol{\nu} \rVert_2^2 &\leq  \lambda \alpha\lVert   \boldsymbol{\beta}^*  \rVert_1- \lambda \alpha \lVert \hat{\boldsymbol{\beta}} \rVert_1+2 \lambda (1-\alpha)\lVert \boldsymbol{L}\hat{\boldsymbol{\beta}} \lVert_2 \, \lVert  \boldsymbol{\beta}^*-\hat{\boldsymbol{\beta}}\rVert_2 -\langle\nabla \ell^{(n)}(\boldsymbol{\theta}^*),\boldsymbol{\nu} \rangle + \tau\, \sqrt{\frac{\log n(p+1)}{n}} \, \lVert \boldsymbol{\nu} \rVert_1^2\\
			&\leq \lambda \alpha \lVert   \boldsymbol{\beta}^*  \rVert_1-  \lambda \alpha \lVert \hat{\boldsymbol{\beta}} \rVert_1+2 \lambda (1-\alpha) \lambda_{max}(\boldsymbol{L}) \lVert  \hat{\boldsymbol{\beta}} \rVert_2 \, \lVert  \boldsymbol{\beta}^*-\hat{\boldsymbol{\beta}}\rVert_2 + \lvert \langle\nabla \ell^{(n)}(\boldsymbol{\theta}^*),\boldsymbol{\nu} \rangle  \rvert + \tau\, \sqrt{\frac{\log n(p+1)}{n}} \, \lVert \boldsymbol{\nu} \rVert_1^2\\
			&\leq   \lambda \alpha \lVert   \boldsymbol{\beta}^*  \rVert_1- \lambda \alpha  \lVert \hat{\boldsymbol{\beta}} \rVert_1+2 \lambda (1-\alpha) \lambda_{max}(\boldsymbol{L}) \, R \, \lVert \boldsymbol{\beta}^*-\hat{\boldsymbol{\beta}}\rVert_2 + \lVert \nabla \ell^{(n)}(\boldsymbol{\theta}^*) \rVert_{\infty} \lVert \boldsymbol{\nu}  \rVert_1 + \tau\, \sqrt{\frac{\log n(p+1)}{n}} \, \lVert \boldsymbol{\nu} \rVert_1^2,\\
				&\leq \lambda \alpha \lVert   \boldsymbol{\beta}^*  \rVert_1- \lambda \alpha \lVert \hat{\boldsymbol{\beta}} \rVert_1+2 \lambda (1-\alpha)\lambda_{max}(\boldsymbol{L}) \, R \, \lVert \boldsymbol{\nu} \rVert_2 + \lVert \boldsymbol{\nu} \rVert_1\, \Bigg[\lVert \nabla \ell^{(n)}(\boldsymbol{\theta}^*)\rVert_{\infty}+\tau  \, R\, \sqrt{\frac{\log n(p+1)}{n}} \Bigg],
		\end{aligned}
		\label{eq:proof7}
	\end{equation*}
where $\lambda_{max}(L)$ is the maximum eigenvalue of $\boldsymbol{L}$. Now using Lemma \ref{lemma2} 
	\begin{equation}	\label{eq:proof9}
		\begin{aligned}
			\gamma \lVert \boldsymbol{\nu} \rVert_2^2 & \leq   \lambda \alpha  \lVert   \boldsymbol{\beta}^*  \rVert_1-  \lambda \alpha \lVert \hat{\boldsymbol{\beta}} \rVert_1+2 \lambda (1-\alpha)\lambda_{max}(\boldsymbol{L}) \, R \, \lVert \boldsymbol{\nu} \rVert_2 +\frac{\lambda \alpha}{2}\lVert \boldsymbol{\boldsymbol{\nu}} \rVert_1\\
			 &  \leq  \frac{3}{2} \lambda \alpha \lVert \boldsymbol{\beta}^* \rVert_1 - \frac{\lambda \alpha}{2}\lVert \hat{\boldsymbol{\beta}} \rVert_1+\frac{\lambda \alpha}{2} \lvert \hat{\sigma} - \sigma^* \rvert + 2 \lambda (1-\alpha)\lambda_{max}(\boldsymbol{L}) \, R \, \lVert \boldsymbol{\nu} \rVert_2.
	\end{aligned}
	\end{equation}
	In case $ \frac{3}{2} \lambda \alpha \lVert \boldsymbol{\beta}^* \rVert_1 - \frac{\lambda \alpha}{2}\lVert \hat{\boldsymbol{\beta}} \rVert_1 > 0$, by using Lemma 5 in \cite{Loh&Wain2015} with $\xi=3$, see Appendix A.1, we obtain
	\begin{equation*}
		\frac{3}{2} \lambda \alpha \lVert \boldsymbol{\beta}^* \rVert_1 - \frac{\lambda \alpha}{2}\lVert \hat{\boldsymbol{\beta}} \rVert_1 \leq \frac{\lambda \alpha}{2} \Bigg(3 \lVert (\hat{\boldsymbol{\beta}}-\boldsymbol{\beta}^*)_A \rVert_1 -  \lVert (\hat{\boldsymbol{\beta}}-\boldsymbol{\beta}^*)_{A^c} \rVert_1\Bigg), 
		\label{eq:proof11}
	\end{equation*}
where,  $A$ is the set of the $s$ largest elements of $\hat{\boldsymbol{\beta}}-\boldsymbol{\beta}^*$	in magnitude,  and $A_c$ is the complementary set, $s$ being defined in Eq.\ \ref{eq:boundedpar}.
	
Then, we have
	\begin{equation*}
		\begin{aligned}
			\gamma \lVert \boldsymbol{\nu} \rVert_2^2 &\leq \frac{\lambda \alpha}{2} \Bigg[ 3 \lVert (\hat{\boldsymbol{\beta}}-\boldsymbol{\beta}^*)_A \rVert_1 - \lVert (\hat{\boldsymbol{\beta}}-\boldsymbol{\beta}^*)_{A^c} \rVert_1,\Bigg] +\frac{3}{2} \lambda \alpha \lvert \hat{\sigma} - \sigma^* \rvert + 2 \lambda (1-\alpha)\lambda_{max}(\boldsymbol{L}) \, R \, \lVert \boldsymbol{\nu} \rVert_2\\
			& \leq \frac{3}{2}\lambda \alpha \lVert \hat{(\boldsymbol{\beta}}-\boldsymbol{\beta}^*)_A \rVert_1 + \frac{3}{2}\lambda \alpha  \lvert \hat{\sigma} - \sigma^* \rvert + 2 \lambda (1-\alpha)\lambda_{max}(\boldsymbol{L}) \, R \, \lVert \boldsymbol{\nu} \rVert_2\\
			& \leq \frac{3}{2}\lambda \alpha \,  \lVert \boldsymbol{\nu}_A\rVert_1+ 2 \lambda (1-\alpha)\lambda_{max}(\boldsymbol{L}) \, R \, \lVert \boldsymbol{\nu} \rVert_2\\
			& \leq \frac{3}{2}\lambda \alpha \sqrt{s+1} \, \lVert \boldsymbol{\nu} \rVert_2+ 2 \lambda (1-\alpha)\lambda_{max}(\boldsymbol{L}) \, R \, \lVert \boldsymbol{\nu} \rVert_2
		\end{aligned}
		\label{eq:proof12}
	\end{equation*}
	from which we finally obtain
	\begin{equation*}
		\begin{aligned}
			\lVert \boldsymbol{\nu} \rVert_2 &\leq \frac{\frac{3}{2}\lambda \alpha \sqrt{s+1} + 2 \lambda (1-\alpha)\lambda_{max}(\boldsymbol{L}) \, R}{\gamma}.
		\end{aligned}
		\label{eq:proof13}
	\end{equation*}

In case   $ \frac{3}{2} \lambda \alpha \lVert \boldsymbol{\beta}^* \rVert_1 - \frac{\lambda \alpha}{2}\lVert \hat{\boldsymbol{\beta}} \rVert_1 \leq 0$, the bound follows easily from Eq.\,(\ref{eq:proof9}).	%
\end{proof}

\section{Numerical implementation} \label{section:opt}

We propose a two-step procedure to numerically solve Eq.\,(\ref{eq: penalized MLE definition}) in AFTNet. In the first step, we estimate $\hat{\sigma}$ using {\tt survreg} function in the {\tt R} package {\tt survival}, i.e., regressing $Y$ only on the intercept. In the second step, we plug this estimate into  Eq.\,(\ref{eq: penalized MLE definition}) and solve the following
\begin{equation}
	\underset{\boldsymbol{\beta}\in \mathbb{R}^p}{\operatorname{argmin}} ~ -\frac{1}{n}\ell(\boldsymbol{\beta})
	+
	\lambda\alpha\sum_{i=1}^{p}\left|\beta_i\right|+\lambda(1-\alpha)\boldsymbol{\beta}^T\mathbf{L}\boldsymbol{\beta}.
	\label{eq: penalized MLE definition gamma}
\end{equation}
We use a proximal gradient technique to obtain the corresponding stationary point of Eq.\,(\ref{eq: penalized MLE definition gamma}). In the following, we give a brief description of this technique. Then, we specialize it for our problem. The proximal gradient technique is an iterative method that solves the composite model  \citep{parikh2014proximal}
\begin{eqnarray*}
	\underset{\boldsymbol{\beta} \in \mathbb{R}^p}{\operatorname{min}} ~ f(\boldsymbol{\beta})+ g(\boldsymbol{\beta}),
\end{eqnarray*}
under the following assumptions (\cite{Beck2017}, ch. 2 and 10)
\begin{itemize}
	\item[(i)] $f$ is proper, closed, differentiable, with Lipschitz-continuous gradient, with convex domain;
	\item[(ii)] $g$ is a proper, closed, and convex function.
\end{itemize}
At each iteration $k$, we linearize the function $f(\boldsymbol{\beta})$ around the current point, and solve the problem
\begin{eqnarray*}
	\underset{\boldsymbol{\beta} \in \mathbb{R}^p}{\operatorname{min}} ~f\left(\boldsymbol{\beta^k}\right)+\nabla f\left(\boldsymbol{\beta^k}\right)^T\left(\boldsymbol{\beta}-\boldsymbol{\beta^k}\right)+ {g}\left(\boldsymbol{\beta}\right)+\frac M 2 \| \boldsymbol{\beta}-\boldsymbol{\beta^k} \|^2_2.
\end{eqnarray*}
The last term is the proximal term whose function is to keep the update in a neighborhood of the current iterate 
$\boldsymbol{\beta^k}$, where $f\left(\boldsymbol{\beta}\right)$ is close to its linear approximation. The constant $M>0$ is the step parameter, an upper bound of the Lipschitz constant of the $\nabla f$.  We can rewrite the problem as
\begin{eqnarray*}
	\underset{\boldsymbol{\beta}\in \mathbb{R}^p}{\operatorname{min}} ~ \frac 1 2 \left\| \boldsymbol{\beta}-\left( \boldsymbol{\beta}^k -\frac 1 M \nabla f\left(\boldsymbol{\beta}^k\right)\right)\right\|^2_2+\frac 1 M {g}\left(\boldsymbol{\beta}\right)
\end{eqnarray*}
whose solution is
\begin{eqnarray*}
	\boldsymbol{\beta}^{k+1}=\mathrm{Prox}_{\frac 1 M {g}\left(\boldsymbol{\beta}\right)}\left( \boldsymbol{\beta}^{k}-\frac 1M \nabla f\left(\boldsymbol{\beta}^{k}\right)\right),
\end{eqnarray*}
where $\mathrm{Prox}_{\frac 1 M {g}\left(\boldsymbol{\beta}\right)}()$ denotes the proximal operator associated to the $\frac 1 M {g}()$ function.

We can now specify such a technique for our case, defining
\begin{eqnarray*}
	f(\boldsymbol{\beta})=-\frac{1}{n}\ell\left(\boldsymbol{\beta}\right)+\lambda(1-\alpha)\boldsymbol{\beta}^T\mathbf{L}\boldsymbol{\beta}=-\frac{1}{n}\ell\left(\boldsymbol{\beta}\right)+\lambda(1-\alpha)\left\|\mathbf{L}^{\frac{1}{2}}\boldsymbol{\beta} \right\|_2^2,
\end{eqnarray*}
and
\begin{eqnarray*}
	g\left(\boldsymbol{\beta}\right)=\lambda\alpha \sum_{i=1}^{p} \left| \beta_i \right|=\lambda\alpha\| \boldsymbol{\beta} \|_1.
\end{eqnarray*}
We recall that the proximal operator of the Lasso penalty $g\left(\boldsymbol{\beta}\right)$ is the soft thresholding.  Hence, we have 
\begin{eqnarray*}
	\boldsymbol{\beta}^{k+1}=\mathrm{SOFT}\left( \boldsymbol{\beta}^{k}-\frac 1M \nabla f\left(\boldsymbol{\beta}^{k}\right),\frac{\lambda\alpha}M\right)
\end{eqnarray*}
which is applied component-wise, i.e.,
\begin{eqnarray*}
	\beta_j^{k+1}=\mathrm{SOFT}\left(u_j,\frac{\lambda\alpha}M\right)=\begin{cases}
		u_j-\frac{\lambda\alpha}{M}, & \text{if}\ u_j>\frac{\lambda\alpha}{M}\\
		0, & \text{if} -\frac{\lambda\alpha}{M} \leq u_j \leq \frac{\lambda\alpha}{M}\\
		u_j+\frac{\lambda\alpha}{M}, & \text{if}\ u_j<-\frac{\lambda\alpha}{M}      
	\end{cases}
	\label{eq:SOFT}
\end{eqnarray*}
with 
\begin{equation*}
	u_j=\beta_j^{k}-\frac 1M \left(\nabla \ell\left(\boldsymbol{\beta}^{k}\right)\right)_j-\frac {2\lambda(1-\alpha)}M \left(\mathbf{L}\boldsymbol{\beta}^k\right)_j,
\end{equation*}
for $j=1,\dots,p$. 

We summarize in Algorithm  \ref{alg:1} the numerical procedure proposed in AFTNet.

\begin{algorithm}[ht!]
	\caption{Proximal gradient algorithm for solving (\ref{eq: penalized MLE definition gamma})}\label{alg:1}
	\begin{algorithmic}
		\Require $ \lambda > 0$
		\State Input:  $\boldsymbol{y}$, $\boldsymbol{X
		}$, $\boldsymbol{\delta}$, $\hat{\sigma}$
		\State  Initialize $k \gets 0 $
		\State  Initialize $\boldsymbol{\beta}^{(k)} $
		\State  Initialize $M >0$
		\While{convergence not satisfied}
		\State {\it Backtracking line search} for the step parameter $M$
		\State evaluate $\nabla \ell\left(\boldsymbol{\beta}^{k}\right)$  by Eq. (\ref{eq:part}) first formula 
		\State  $\boldsymbol{u} \gets \boldsymbol{\beta}^{k}-\frac 1 M \left(\nabla \ell \left(\boldsymbol{\beta}^{k}\right)\right)-\frac {2\lambda(1-\alpha)}M \left(\mathbf{L}\boldsymbol{\beta}^k\right)$
		\State $\boldsymbol{\beta}^{k+1}  \gets \mathrm{SOFT}\left( \boldsymbol{u},\frac{\lambda\alpha}M\right)$
		\State {\it Check convergence}
		\State  $k \gets k+1$
		\EndWhile
	\end{algorithmic}
\end{algorithm}
In Algorithm \ref{alg:1}, {\it Backtracking line search} for the step parameter $M$ is implemented as in \cite{Beck2017}.  
The {\it Check convergence} stops the algorithm when a maximum number of iterations has been reached, or the relative error norm between the estimates at two consecutive iterations is less than a given tolerance. 
Furthermore, Algorithm \ref{alg:1} assumes that the regularization parameter $\lambda$ is known. In practice, we need to choose the regularization parameters by some data-driven selection procedure; in the following subsection, we describe our choice.

We implemented the proposed algorithm in the R package AFTNet, along with a set of auxiliary functions for fitting,  cross-validation, prediction and visualization. The code is available upon request. The R version used is 4.3.0 (2023-04-21).
 
\subsection{$\lambda$ parameter selection}\label{section: cross}


In our AFTNet implementation, we extend the cross-validated linear predictors approach (CV-PL)  \cite{dai2019cross}  to the Weibull AFT penalized model. More precisely, after dividing the training data into $K$ non-overlapping folds,  for each $k=1, \ldots, K$, we use the $k$-th fold $D^k$ as test set, and the $K-1$ remaining folds $T^k$ as the training set. We obtain $\boldsymbol{\hat{\beta}}_{\lambda}^{(-k)}$ from $T^k$. Then, we evaluate the standardized residuals
\begin{equation*}
	 \hat{e}^{CV}_{i,\lambda}=\frac{y_i-\boldsymbol{x}_i^\top \boldsymbol{\hat{\beta}}_{\lambda}^{(-k)}}{\hat{\sigma}}, \quad \forall i\in D^k.
	\label{eq1}
\end{equation*}
Hence,  after repeating this for all K folds,  we combine all standardized residuals $\hat{e}^{CV}_{i,\lambda} \quad i=1,\ldots ,n$.  We plug in $\hat{e}^{CV}_{i,\lambda}$ into the negative log-likelihood in Eq. \eqref{eq: loglike per weibull} and define $CV(\lambda)$ as  
\begin{equation*}
		CV(\lambda)= -\sum_{i=1}^n \delta_i \left( - \log \big(\hat{\sigma}\big) + \hat{e}_{i,\lambda}^{CV}\right) - \exp \left( \hat{e}_{i,\lambda}^{CV}\right).
\end{equation*}

We evaluate $CV(\lambda)$ at each value of $\lambda$ belonging to a predetermined grid of values within the interval $[\lambda_{min}, \lambda_{max}]$ and select the one for which $CV(\lambda)$ is the minimum.

\section{Simulation results}\label{section: simul}
 To demonstrate the performance of AFTNet, we perform various numerical experiments on synthetic data sets and compare AFTNet with the elastic net regularized Gehan estimator under the AFT model described in \cite{suder2022scalable}. 
  In each synthetic dataset, we split $n$ observations into $n_T$ observations assigned to the training set and $n_D$  observations to the test set, with $n=n_T+n_D$. The latter serves to evaluate the prediction performance.  
We consider scenarios that are likely to be encountered in genomic studies for biomarkers discovery, i.e., number of genes/variables $p >> n_T$ and availability of prior information about the regulatory networks among genes encoded by a known adjacency matrix $\mathbf{A}$. In all our experiments, the network consists of $r$ regulatory modules. Similarly to \cite{sun2014network},  
  each $i$-th module, $i=1,\dots,r$, models one transcription factor (TF) that regulates a given number of genes $p_i$, such that the overall number of genes is  $p=\sum_{i=1}^r (p_i+1)$.
 We define the adjacency matrix $\mathbf{A}$ as a binary matrix of dimension $p\times p$, with $a_{ij} = 1$ between the TFs and their regulated genes, and $a_{ij} = 0$ otherwise. We consider two cases $r = 20$ and $r=100$ according to Table \ref{tab1}, where $p/n_T=2$ represents a \emph{weak} effect of high-dimensionality and $p/n_T=4$  represents a \emph{strong} effect of high-dimensionality.

\begin{table}[htb]
	\begin{center}
		\begin{tabular}{ c c c c} 
			\\[-1.8ex]\hline
			\hline \\[-1.8ex] 
			Effect & Number of genes& Training set & Testing set  \\
			& $p$ & $n_{T}$ & $n_{D}$ \\
			\\[-1.8ex]\hline
			\hline \\[-1.8ex] 
			Weak & 220     & 110 & 55 \\
			Strong& 1100 &  275& 138 \\
			\hline
		\end{tabular}
	\end{center}
	\caption{Simulation studies' dimensions: Weak and strong high-dimensionality effects are tested in separate evaluations using both not-overlapping and overlapping cases. $p$ is the number of potential explanatory variables (i.e., genes). $n_{T}$ and $n_{D}$ are the number of observations in the traning set $T$ and testing set $D$, respectively. 
	}
	\label{tab1}
\end{table}

We consider two topological settings:

\begin{itemize}
\item {\bf not-overlapping regulatory modules}:
The $r$ regulatory modules are disjoint from each other. In each module,  the TF regulates 10 genes. For each sample $i$, $i=1,\ldots, n$, the $i$-th row of the matrix $\mathbf{X}$ is given by the expression values of the $p$ genes generated according to the following scheme: the expression value of each $TF_j$, $j=1, \ldots, r$, is sampled from a standard normal distribution. The expression values of the ten regulated genes are sampled from a conditional normal distribution with correlation $\rho$ between their expressions and that of the corresponding $TF$.
For each module, we randomly select $v$ genes to have a positive correlation $\rho = 0.7$ and the remaining $10-v$ genes to have a negative correlation $\rho =- 0.7$, mimicking the activation or repression of each gene under the effect of its corresponding $TF$.

\item {\bf overlapping regulatory modules}:
The first four regulatory modules overlap. In particular, the first two regulatory modules share 10 common genes (i.e., jointly regulated by $TF_1$ and $TF_2$), 5 genes are specific to the first regulatory module, and 5 genes are specific to the second regulatory module. Therefore, the first two regulatory modules contain 20 genes and 2 TFs.
The third and fourth regulatory modules have 6 common genes (i.e., jointly regulated by $TF_3$ and $TF_4$), 7 genes are specific to the third regulatory module, and 7 are specific to the fourth module. Therefore, together, they are composed of 20 genes and 2 TFs. 
The remaining $r-4$ regulatory modules do not overlap and the TF regulates 10 genes in each module, as in the not-overlapping case.
This scenario mimics cases where some genes can belong to different pathways regulating different biological processes, as often observed in cancer.
In this second setting, for each sample $i$, $i=1, \ldots, n$, the $i$-th row of the matrix $\mathbf{X}$ is given by the expression values of the $p$ genes generated according to the following scheme: the expression value of each $TF_j$, $j=1,\ldots, r$, is sampled from a standard normal distribution, the expression values of the module-specific regulated genes are sampled from a conditional normal distribution with correlation $\rho$ between their expressions and that of the corresponding TF. In contrast, the expression values of the common regulated genes are sampled from a conditional normal distribution with correlation $\rho$ between their expressions and that of the average of the two corresponding TFs.
For both module-specific and common genes, we randomly select $v$ genes to have a positive correlation $\rho = 0.7$ and the remaining genes to have a negative correlation $\rho =- 0.7$, mimicking the activation or repression of each gene under the effect of its corresponding $TF$.  
\end{itemize}
\noindent

In both dimensional cases, \emph{weak} and \emph{strong}, and  topological settings, {\bf not-overlapping} and {\bf overlapping}, we consider $p_{active}=88$ active genes, choosing the true parameter vector $\boldsymbol{\beta}^* \in \mathbb{R}^p$ with the last $p-88$ components equal to 0, and  the coefficients $\beta^*_ j, \, j=1, \ldots, 44$ are generated from the uniform distribution $\mathcal{U}(0.1, 0.5)$,  while $\beta^*_ j, \, j=45, \ldots, 88$ are generated from  $\mathcal{U}(-0.5, -0.1)$.
\noindent
In each dimensional case and  setting, we generate times, $T_i$, $i=1, \ldots, n$, sampling from a Weibull distribution with shape parameter equal to $1/\sigma$ and scale parameter equal to $\exp(\boldsymbol{x_i}^T\boldsymbol{\beta}^*)$, with $\boldsymbol{x_i}^T$ being the $i$-th row of matrix $\boldsymbol{X}$. We $\log$-transform the times, and we apply a 30\% censoring rate. 

\noindent
We consider three values for the scale parameter $\sigma$:  $\sigma=0.5$ corresponding to increasing hazard, $\sigma=1$ corresponding to an exponential distribution with constant hazard, and $\sigma=1.5$ corresponding to decreasing hazard.

\noindent
We use AFTNet as described in Section \ref{section:opt}. In particular, in Algorithm \ref{alg:1}, we initialize $\boldsymbol{\beta}^{(0)}=\boldsymbol{0}$, we fix $\alpha=0.5$ and $M$ the largest eigenvalue of the likelihood Hessian matrix evaluated in $(\boldsymbol{\beta}^{(0)},\hat{\sigma})$. We select the optimal parameter $\lambda_{opt} \in [\lambda_{min},\lambda_{max}]$ by the CV-LP approach illustrated in subsection \ref{section: cross} with $K=5$ folds, $\lambda_{max}= \left\| \nabla \ell \left(\boldsymbol{\beta}^{(0)}, \hat{\sigma} \right) \right\|_{\infty} / \alpha$ and $\lambda_{min}=0.01 \cdot \lambda_{max}$. 
For the $\lambda$'s grid we consider 50 equispaced points $\zeta_i$ in the interval $[\log10(\lambda_{min}),\log10(\lambda_{max})]$ and take $\lambda_i=10^{\zeta_i}$.

For what concerns the implementation of the regularized Gehan estimator, we use the {\tt penAFT} package with the following choice penalty=``EN", which stands for elastic-net penalty, with $\alpha=0.5$, $nlambda=50$ different value of regularization parameter and $lambda.ratio.min=0.01$.
%
%
 

\vspace{0.5cm}

\noindent
To compare performance of the different approaches, we use the following indicators:

\begin{itemize}
	\item [-] the Estimated Mean Square Error (EMSE):
	\begin{equation*}
		EMSE=\left\| \boldsymbol{\beta}^*-\hat{\boldsymbol{\beta}}\right\|_2,
		\label{eq4}
	\end{equation*}
	\item [-] the Predictive Mean Square Error (PMSE):
	\begin{equation*}
		PMSE=\left\| \boldsymbol{X}_D^T\, \boldsymbol{\beta}^*-\boldsymbol{X}_D^T\, \hat{\boldsymbol{\beta}}\right\|_2^2  /n_D,
		\label{eq5}
	\end{equation*}
	\item [-] the False Negative Rate (FNR):
	\begin{equation*}
		FNR = \sum_{j=1}^p I \left[ \hat{\boldsymbol{\beta}}_j=0 \land \boldsymbol{\beta}^*_j \neq 0\right] / p_{active},
		\label{eq6}
	\end{equation*}
	which is the relative number of not detected genes,
	\item [-] the False Positive Rate (FPR):
	\begin{equation*}
		FPR = \sum_{j=1}^p I \left[\hat{\boldsymbol{\beta}}_{j}\neq 0 \land \boldsymbol{\beta}^*_j= 0\right] / (p-p_{active}),
	\end{equation*}
	which is the relative number of (falsely) detected genes,
	\item [-] the Number of Selected variables Rate (NSR):
	\begin{equation*}
		NSR = \sum_{j=1}^p I \left[\hat{\boldsymbol{\beta}}_{j} \neq 0\right] / p,
	\end{equation*}
	\end{itemize}
where  $\hat{\boldsymbol{\beta}}$ denotes the estimated vector coefficients obtained by AFTnet or by {\tt penAFT} package.

In the following,  we show the average of the above indicators over 20 independent simulations.

\vspace{0.5cm}

\noindent
{\bf Results for not-overlapping case}

We report simulation results for both models in Tables \ref{tab2} and \ref{tab3} for the weak and strong dimensional cases, respectively.   Across all the considered topological settings and associated choices of $\sigma$ (i.e., $\sigma=0.5, 1, 1.5$), AFTNet outperforms penAFT in both the weak and strong effect in terms of EMSE, PMSE, and FNR. In contrast, penAFT tends to select fewer genes, resulting in a lower FPR.

\begin{figure}[ht]
	\centering
	\includegraphics[scale=0.23]{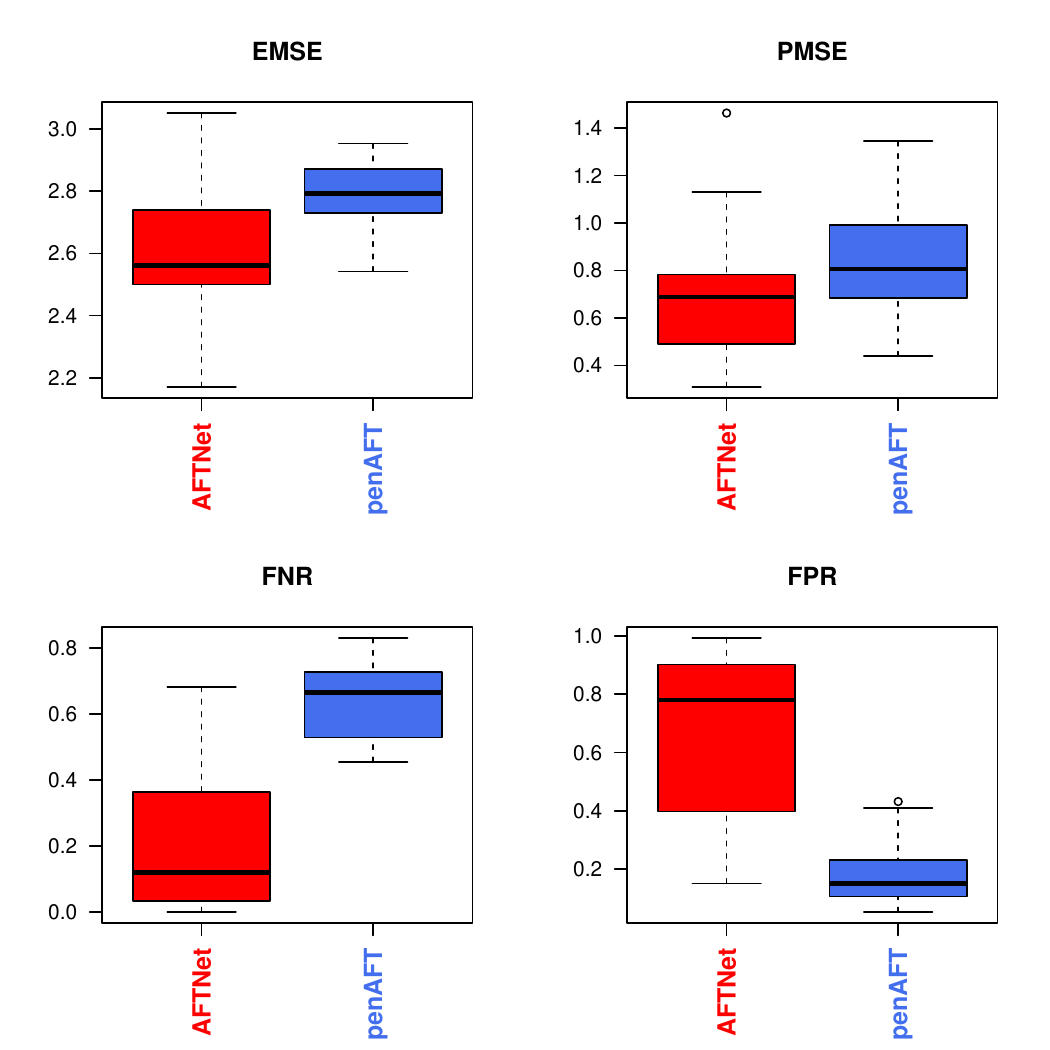}
	\includegraphics[scale=0.23]{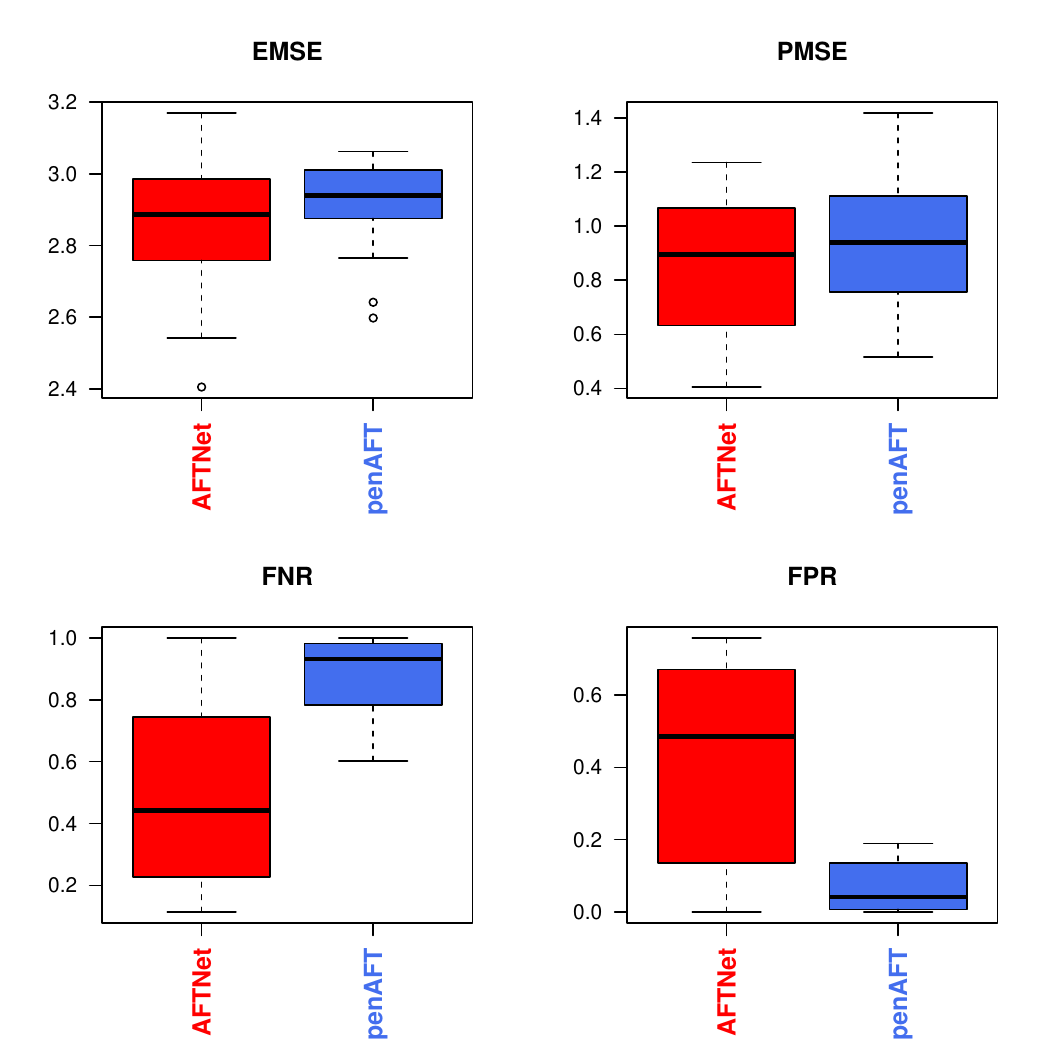}
	\includegraphics[scale=0.23]{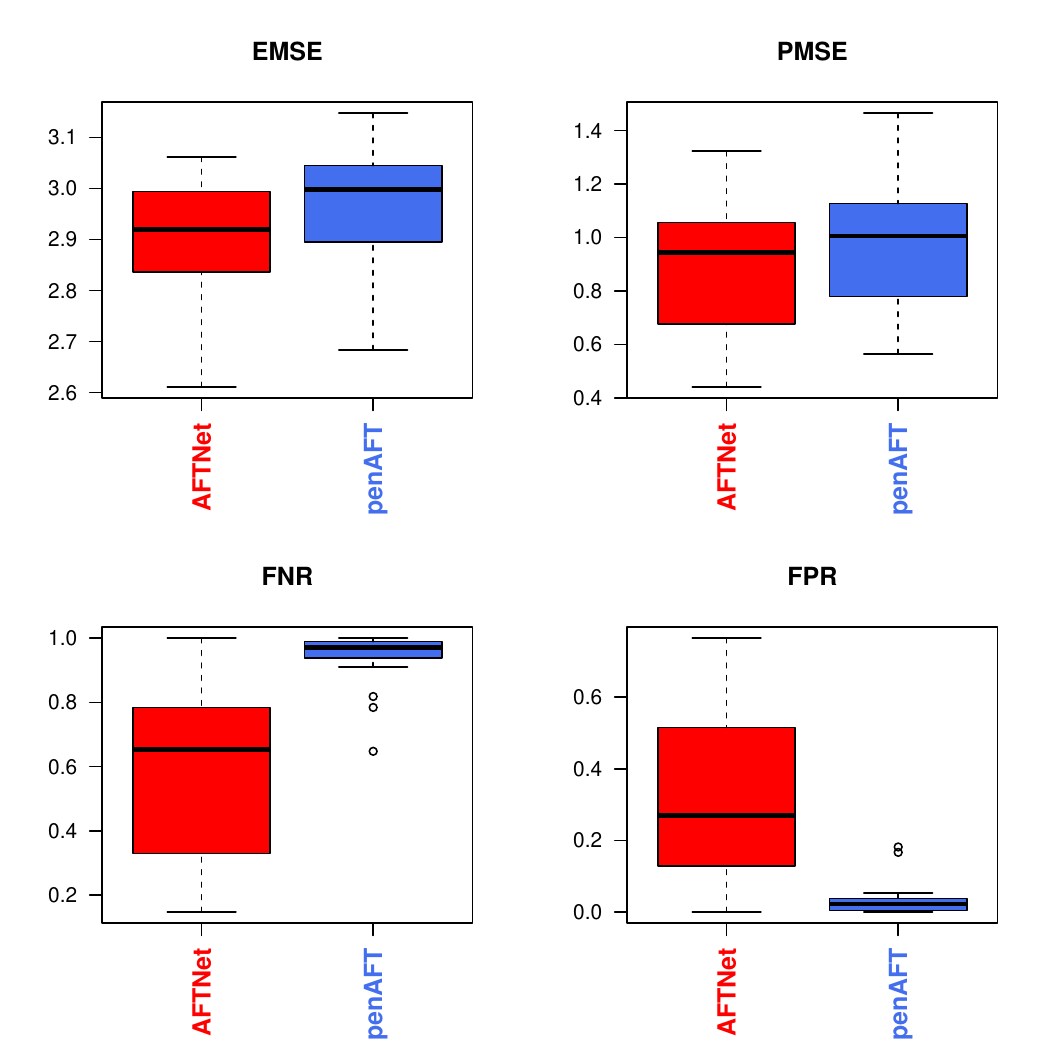}
	\caption{{\bf Not-overlapping case}. Box-plots of the performance metrics results between AFTNet and penAFT with $n_T=110$, $n_D=55$, $p=220$ (weak effect) and $\alpha=0.5$ for $\sigma=0.5, 1, 1.5$ (from the left-side to the right-side), respectively, averaged over 20 independent replications.}
	\label{fig1}
\end{figure}

\begin{figure}[ht]
	\centering
	\includegraphics[scale=0.23]{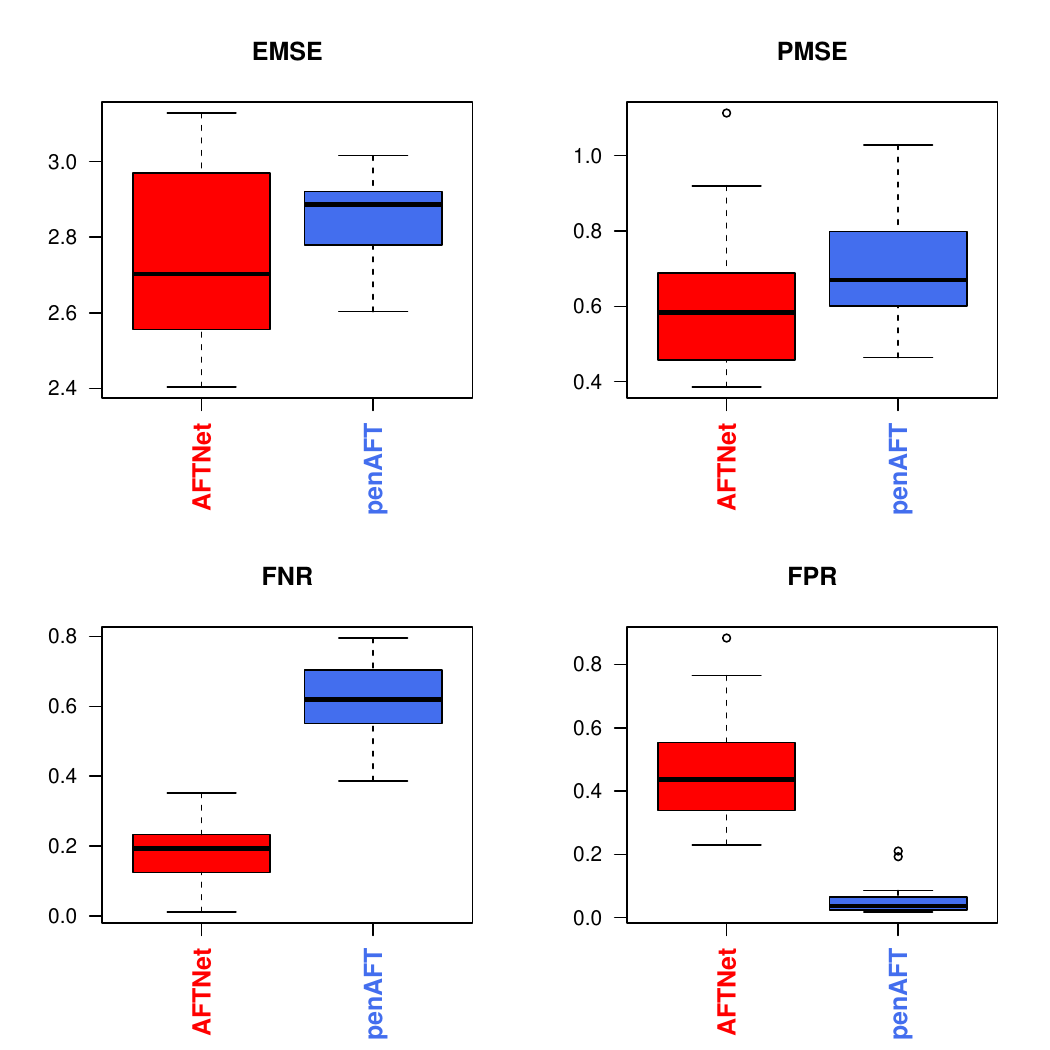}
	\includegraphics[scale=0.23]{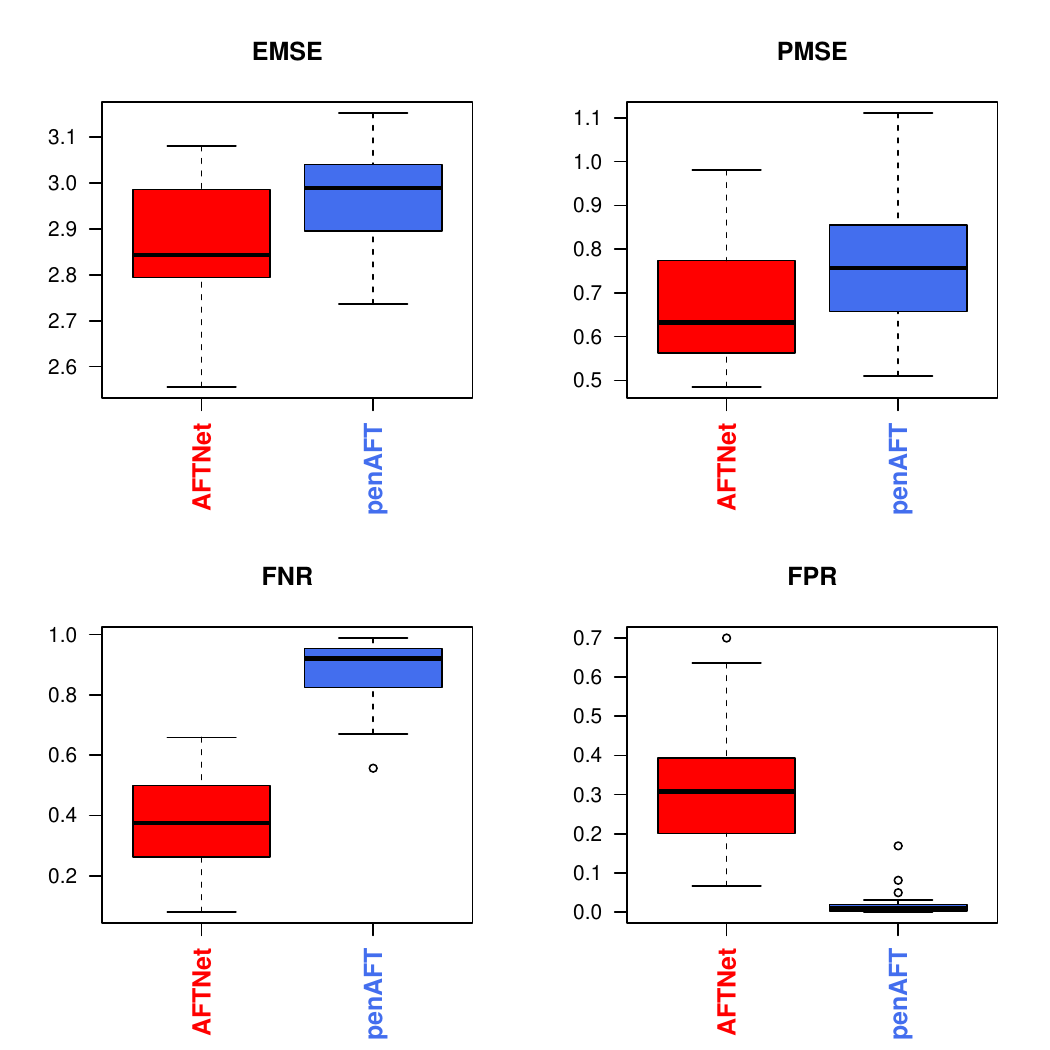}
	\includegraphics[scale=0.23]{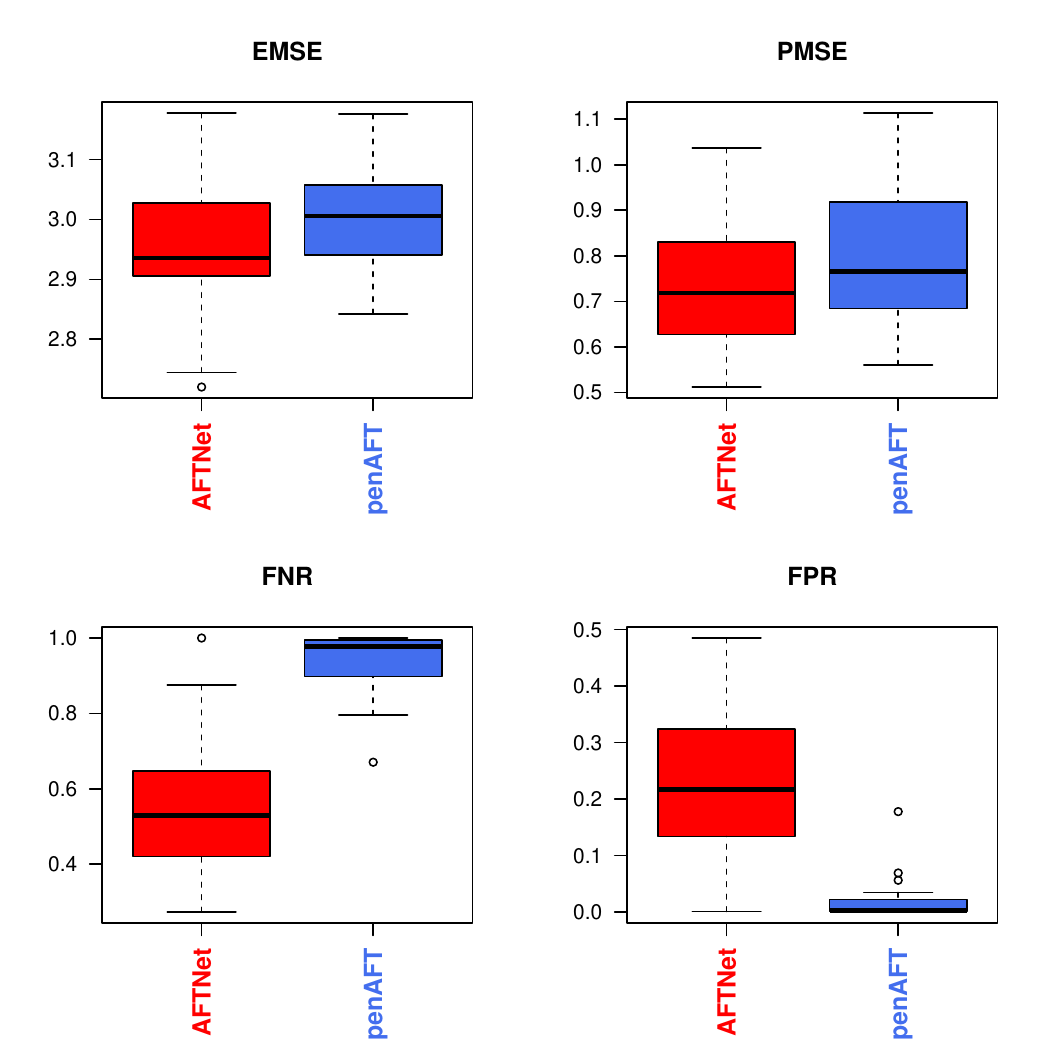}
	\caption{{\bf Not-overlapping case}. Box-plots of the performance metrics results between AFTNet and penAFT with $n_T=275$, $n_D=138$, $p=1100$ (strong effect) and $\alpha=0.5$ for $\sigma=0.5, 1, 1.5$ (from the left-side to the right-side), respectively, averaged over 20 independent replications.}
	\label{fig2}
\end{figure}

\begin{figure}[ht]
	\centering
	\includegraphics[scale=0.35]{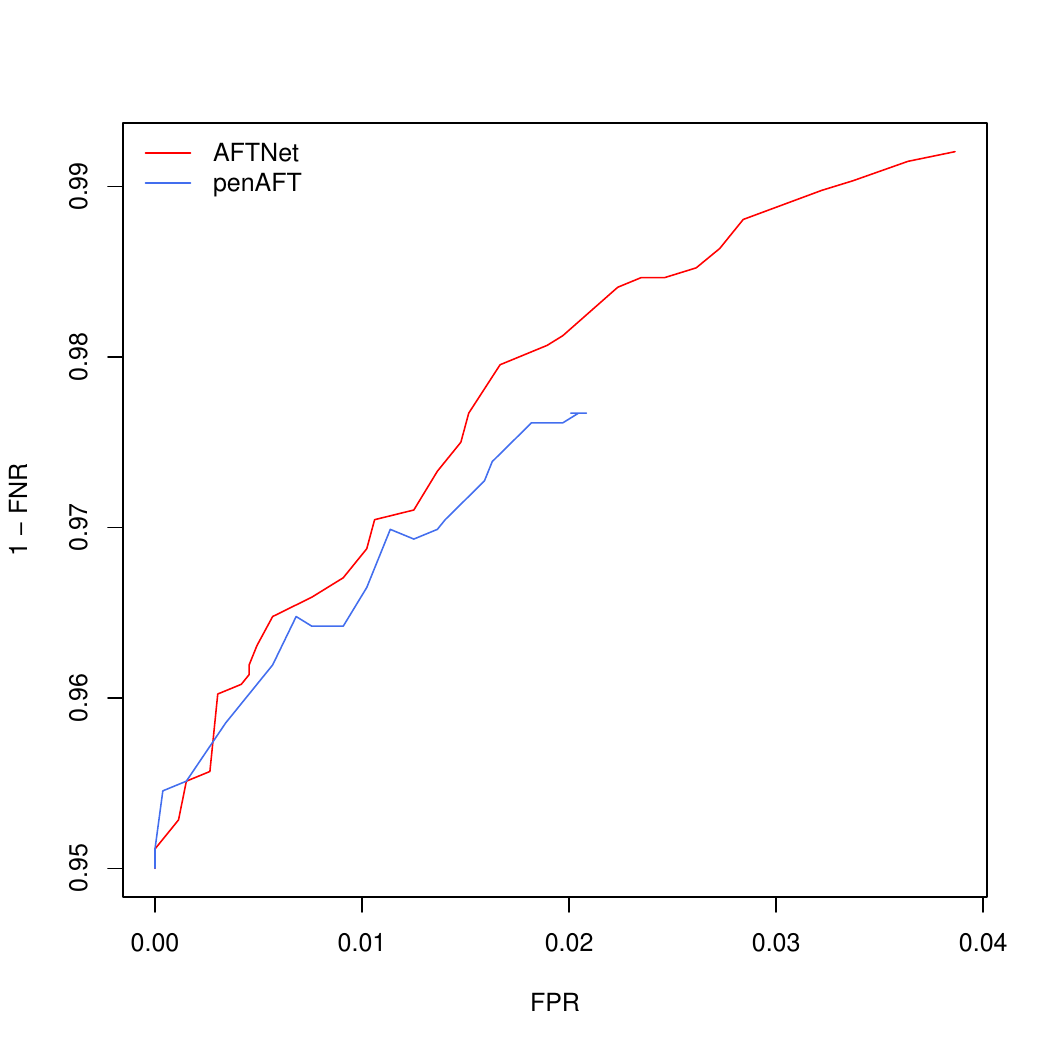}
	\includegraphics[scale=0.35]{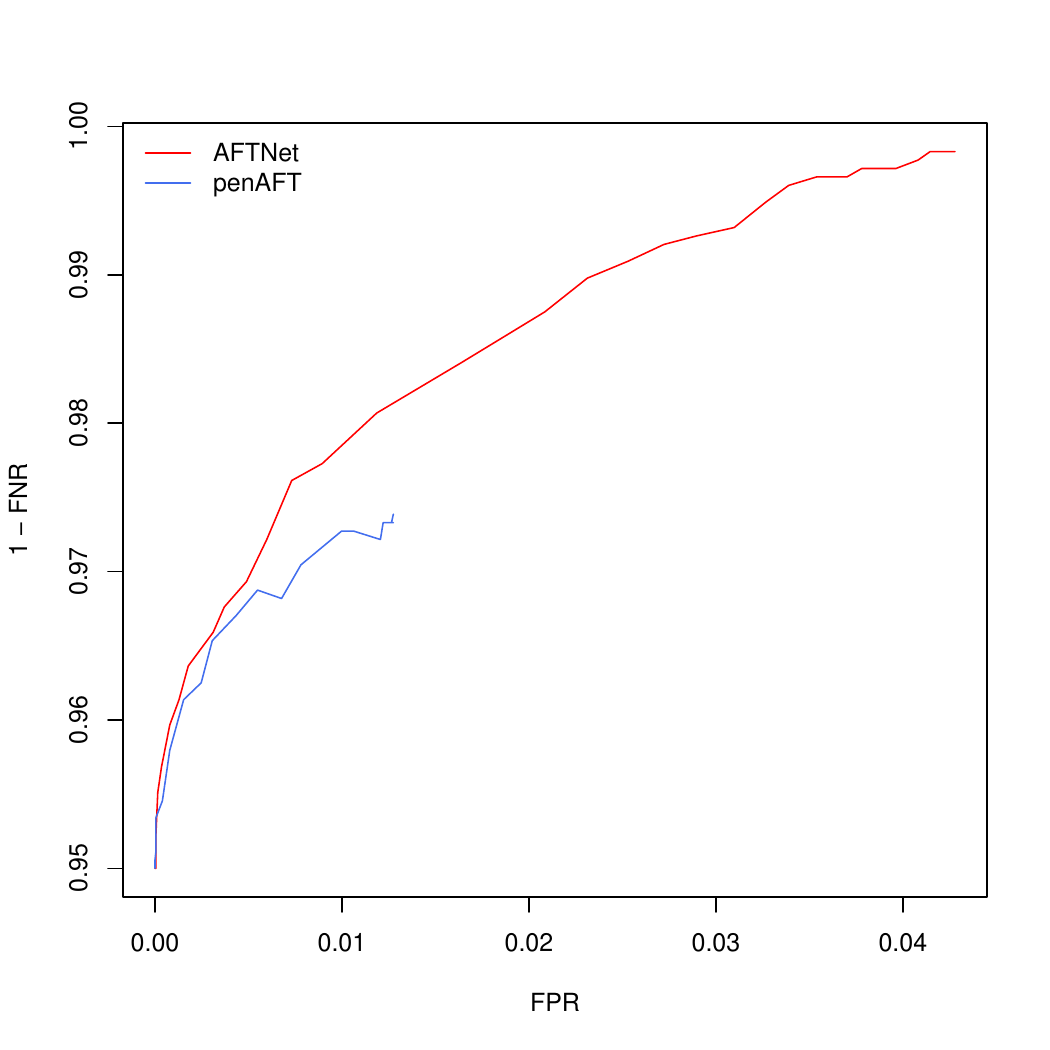}
	\caption{{\bf Not-overlapping case}. ROC curve for $\sigma=1$ with a weak effect (on the left-side) and a strong effect (on the right-side).}
	\label{fig3}
\end{figure}

\begin{table}[!htbp] \centering 
	\caption{{\bf Not-overlapping case}. Performance metrics with $n_T=110$, $n_D=55$, $p=220$ and $\alpha=0.5$ (weak effect) averaged over 20 independent replications.} 
	\label{tab2} 
	\begin{tabular}{@{\extracolsep{5pt}} lcccccc} 
		\\[-1.8ex]\hline
		\hline \\[-1.8ex] 
		& \multicolumn{2}{c}{$\sigma=0.5$}   & \multicolumn{2}{c}{$\sigma=1$}  & \multicolumn{2}{c} {$\sigma=1.5$} \\ 
		\hline \\[-1.8ex] 
		& AFTNet & penAFT & AFTNet& penAFT & AFTNet & penAFT\\ 
		\hline \\[-1.8ex] 
		EMSE & 2.609 & 2.785 & 2.855 & 2.911 & 2.891 &	2.954 \\
		PMSE & 0.685 & 0.825 & 0.846 &	0.928 & 0.892 &	0.962 \\
		FNR    & 0.206 & 0.640 & 0.495 & 0.872 & 0.592 &	0.939 \\
		FPR    & 0.664 & 0.185 & 0.405 & 0.067 & 0.319 &	0.0341 \\
		NSR    & 0.716 & 0.255 & 0.445 & 0.092 & 0.355 &	0.045 \\
		\hline \\[-1.8ex] 
	\end{tabular} 
\end{table} 
\begin{table}[!htbp] \centering 
	\caption{{\bf Not-overlapping case}. Performance metrics with $n_T=275$, $n_D=138$, $p=1100$ and $\alpha=0.5$ (strong effect) averaged over 20 independent replications.} 
	\label{tab3} 
	\begin{tabular}{@{\extracolsep{5pt}} lcccccc} 
		\\[-1.8ex]\hline
		\hline \\[-1.8ex] 
		& \multicolumn{2}{c}{$\sigma=0.5$}   & \multicolumn{2}{c}{$\sigma=1$}  & \multicolumn{2}{c} {$\sigma=1.5$} \\ 
		\hline \\[-1.8ex] 
		& AFTNet & penAFT & AFTNet & penAFT & AFTNet & penAFT\\ 
		\hline \\[-1.8ex] 
		EMSE & 2.731 & 2.851  & 2.854 &	2.968 & 2.949 &	2.999 \\
		PMSE & 0.609 &	0.694 & 0.677 &	0.773 & 0.748 &	0.794 \\
		FNR &  0.183 &	0.619 & 0.374 &	0.874 & 0.558 &	0.935 \\
		FPR &  0.472 &	0.056 & 0.317 &	0.023 &  0.224 &0.021 \\
		NSR &  0.499 &	0.082 & 0.342 &	0.031 & 0.242 &	0.024 \\
		\hline \\[-1.8ex] 
	\end{tabular} 
\end{table} 
Figures \ref{fig1} and  \ref{fig2} provide analogous results using box-plots.  Here,  a greater variability is visible in AFTNet for FNR and FPR indicators.
%
Moreover, to assess the validity of our method, we plot the ROC curve (for  $\sigma=1$) in Figure \ref{fig3}. The figure shows that AFTnet performs better than penAFT in weak and strong dimensional scenarios.

Finally, the average running time of AFTNet for obtaining a single solution path over 20 independent replications, for $\sigma=0.5, 1, 1.5$, in the simulation settings with $(n, p) = (165, 220)$  is about 5.99, 5.98, 6.20 seconds, respectively. Whereas, with  $(n, p) = (413, 1100)$,  is about  538.28, 237.98, 240.06 seconds, respectively. In the first case,  the running time of AFTNet is slightly above the average running time for the penAFT; in the second one, it is larger. However, the comparison is only partially fair since penAFT implementation uses C++ functions to speed up the running time.

\vspace{1cm}

\noindent
{\bf Results for overlapping case}

\begin{figure}[ht]
	\centering
	\includegraphics[scale=0.23]{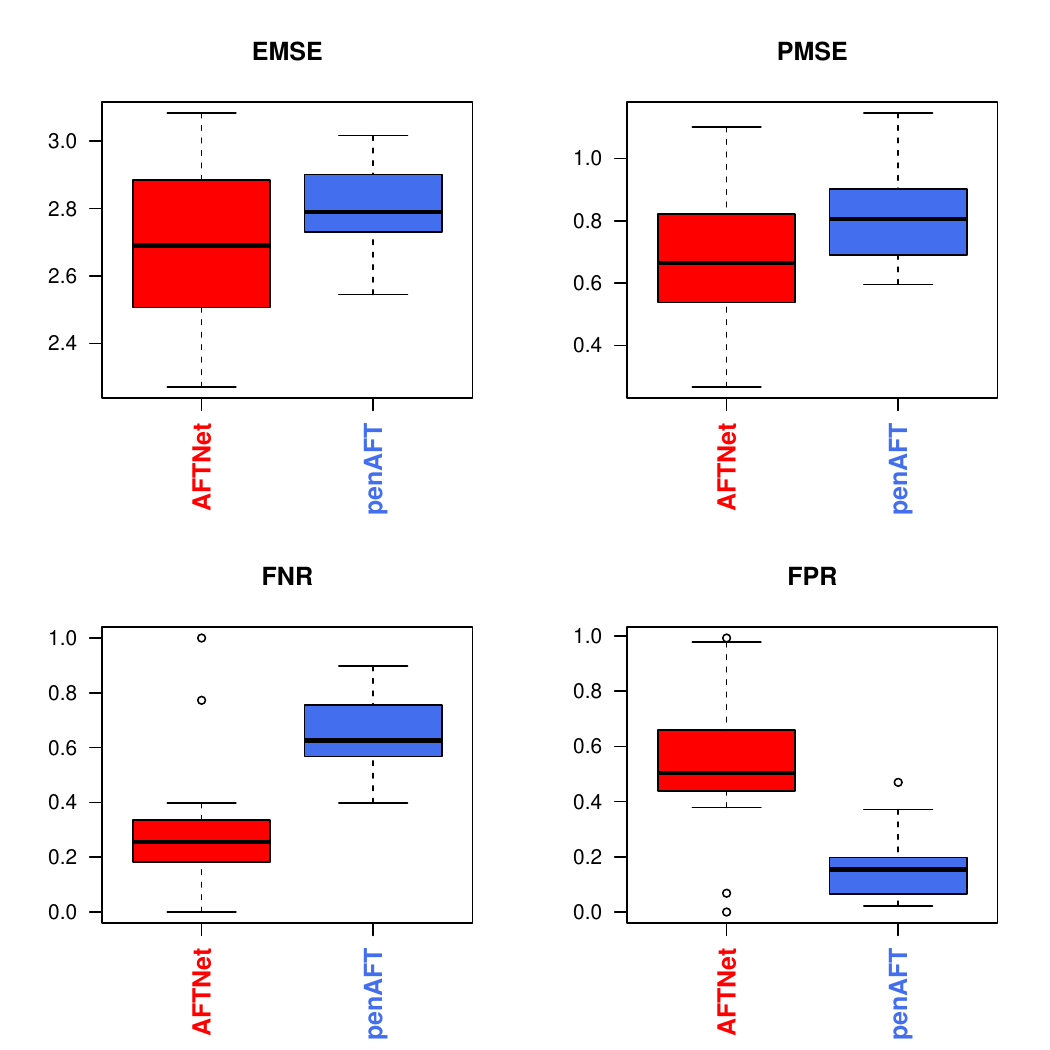}
	\includegraphics[scale=0.23]{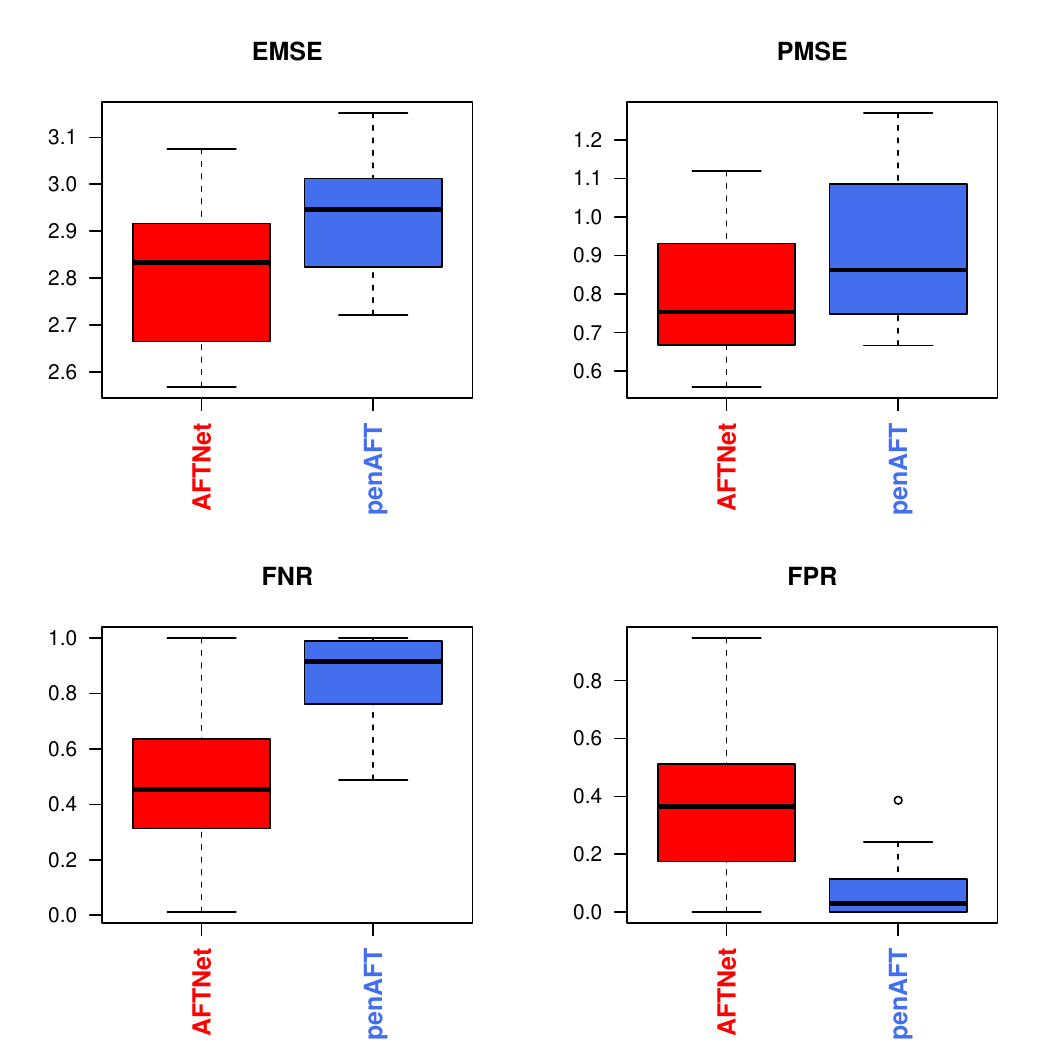}
	\includegraphics[scale=0.23]{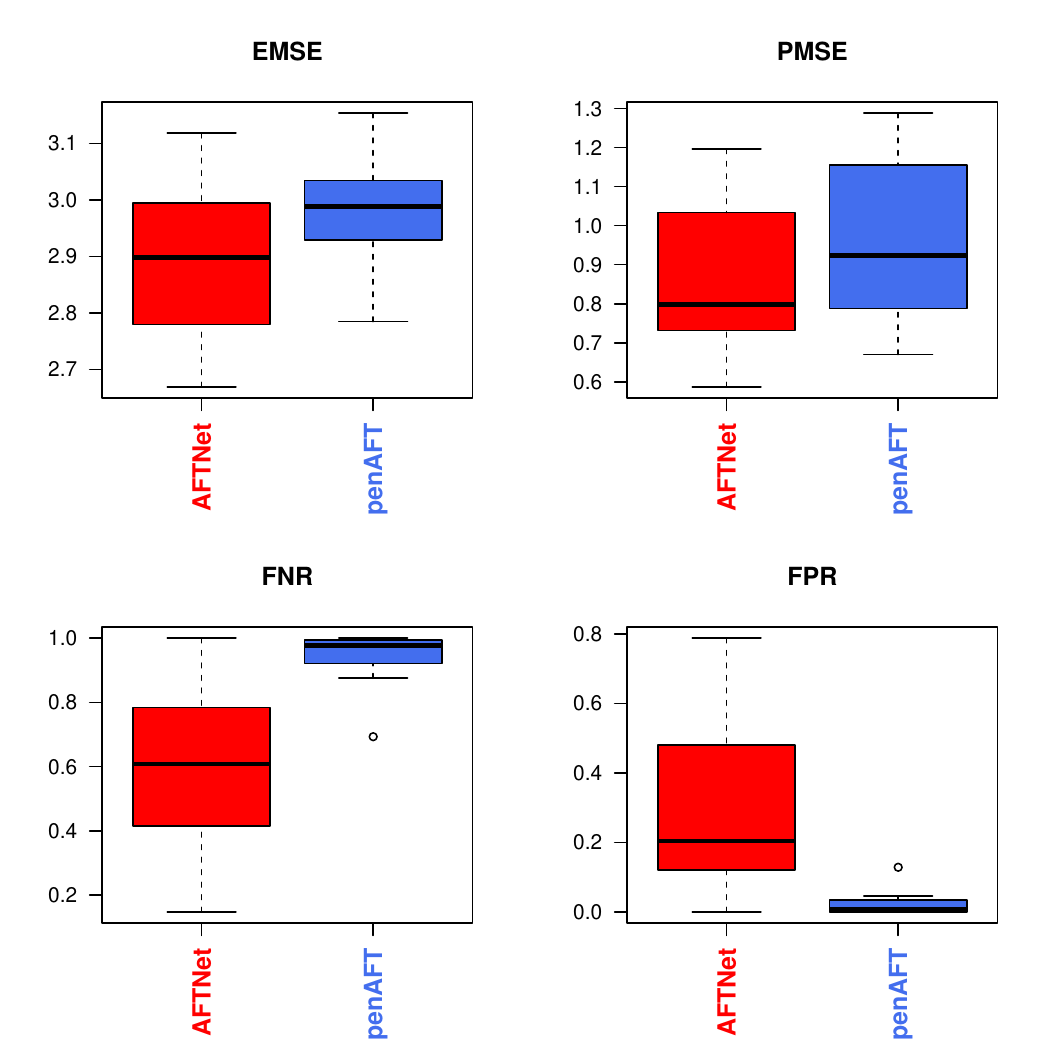}
	\caption{{\bf Overlapping case}. Box-plots of the performance metrics results between AFTNet and penAFT with $n_T=110$, $n_D=55$, $p=220$ (weak effect) and $\alpha=0.5$ for $\sigma=0.5, 1, 1.5$ (from the left-side to the right-side), respectively, averaged over 20 independent replications.}
	\label{fig4}
\end{figure}

\begin{figure}[ht]
	\centering
	\includegraphics[scale=0.23]{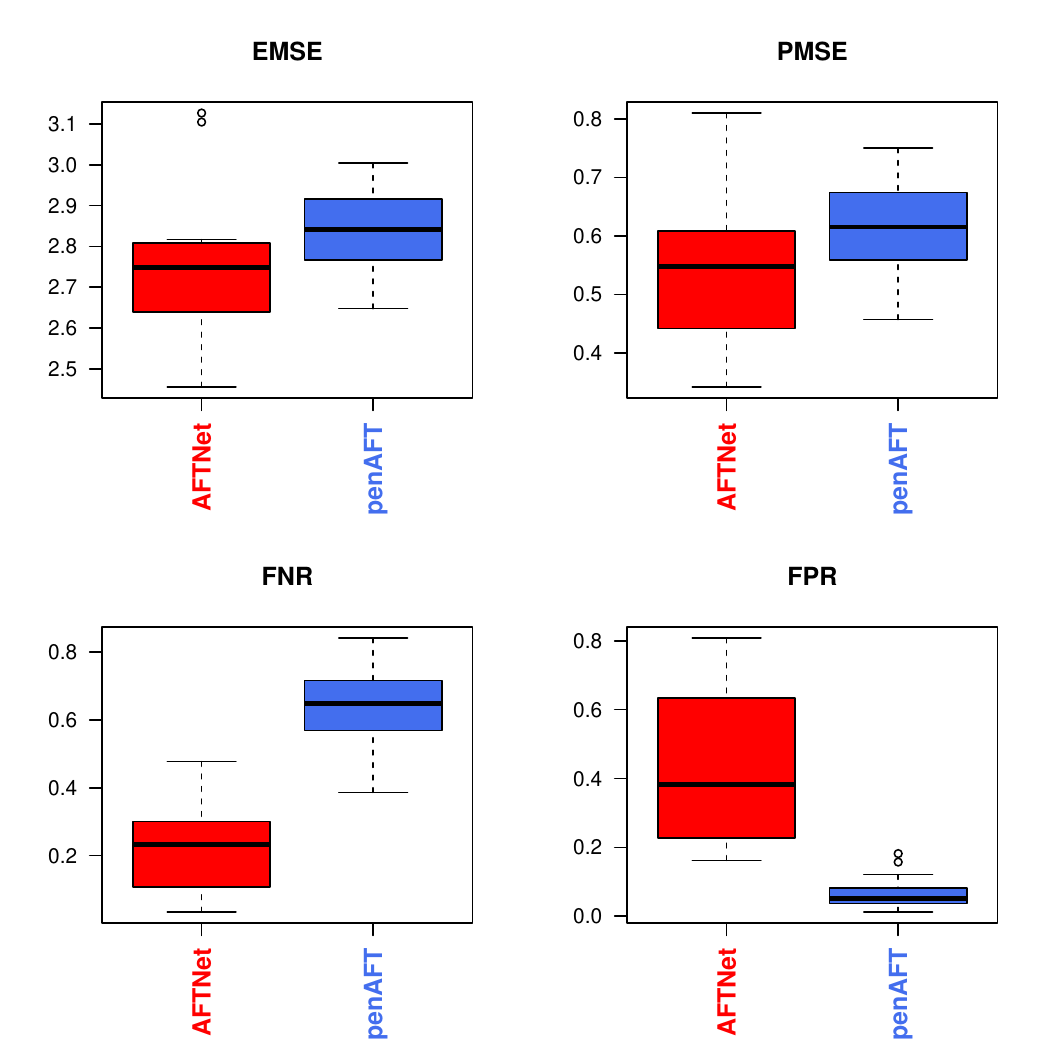}
	\includegraphics[scale=0.23]{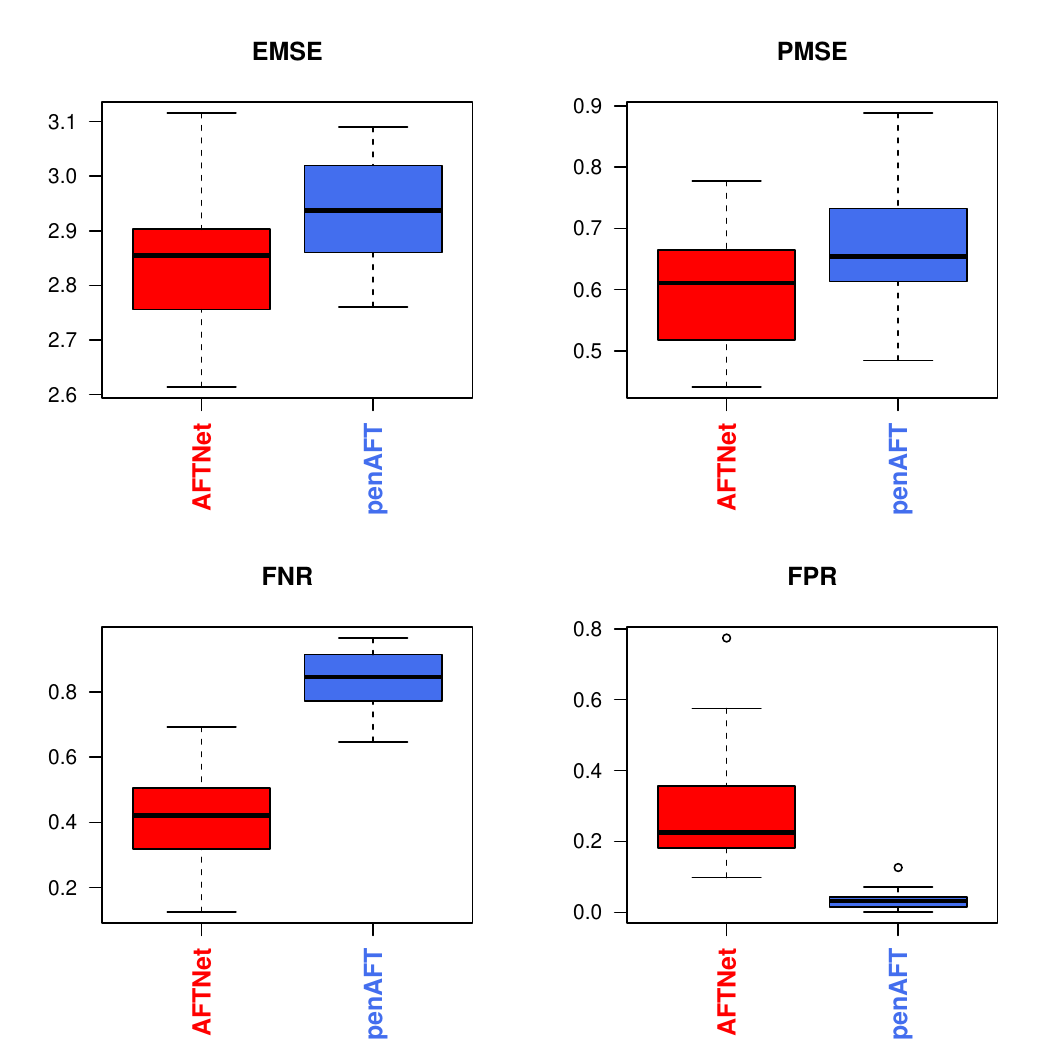}
	\includegraphics[scale=0.23]{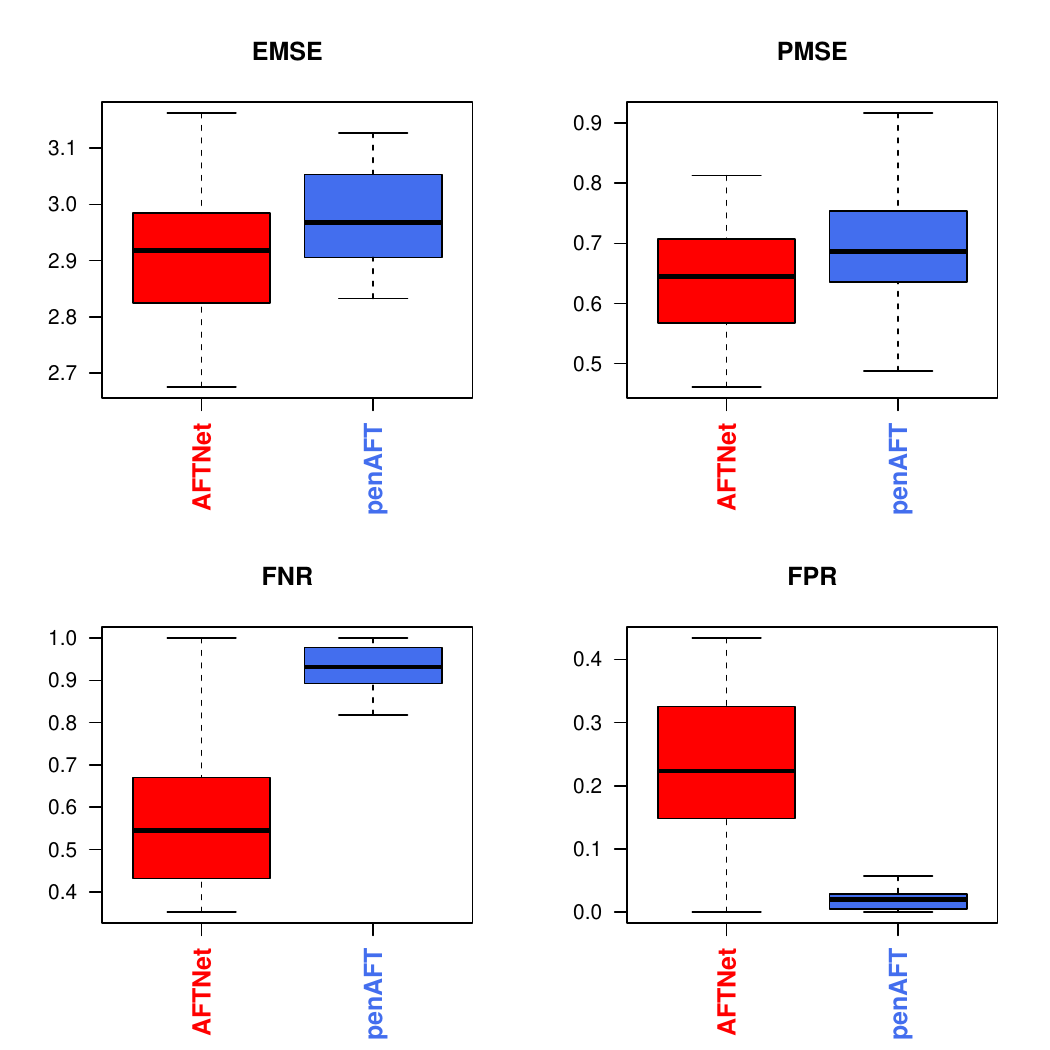}
	\caption{{\bf Overlapping case}. Box-plots of the performance metrics results between AFTNet and penAFT with $n_T=275$, $n_D=138$, $p=1100$ (strong effect) and $\alpha=0.5$ for $\sigma=0.5, 1, 1.5$ (from the left-side to the right-side), respectively, averaged over 20 independent replications.}
	\label{fig5}
\end{figure}

\begin{figure}[ht]
	\centering
	\includegraphics[scale=0.35]{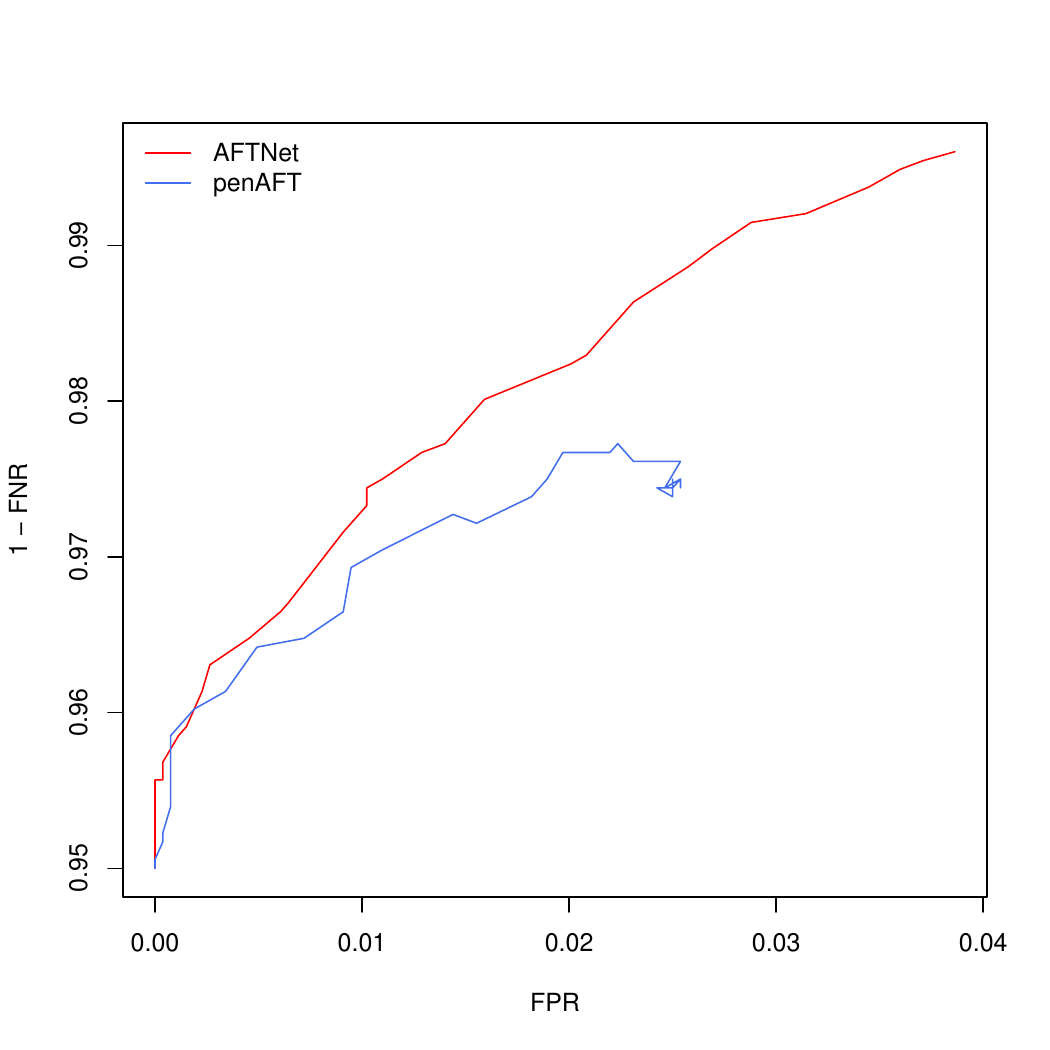}
	\includegraphics[scale=0.35]{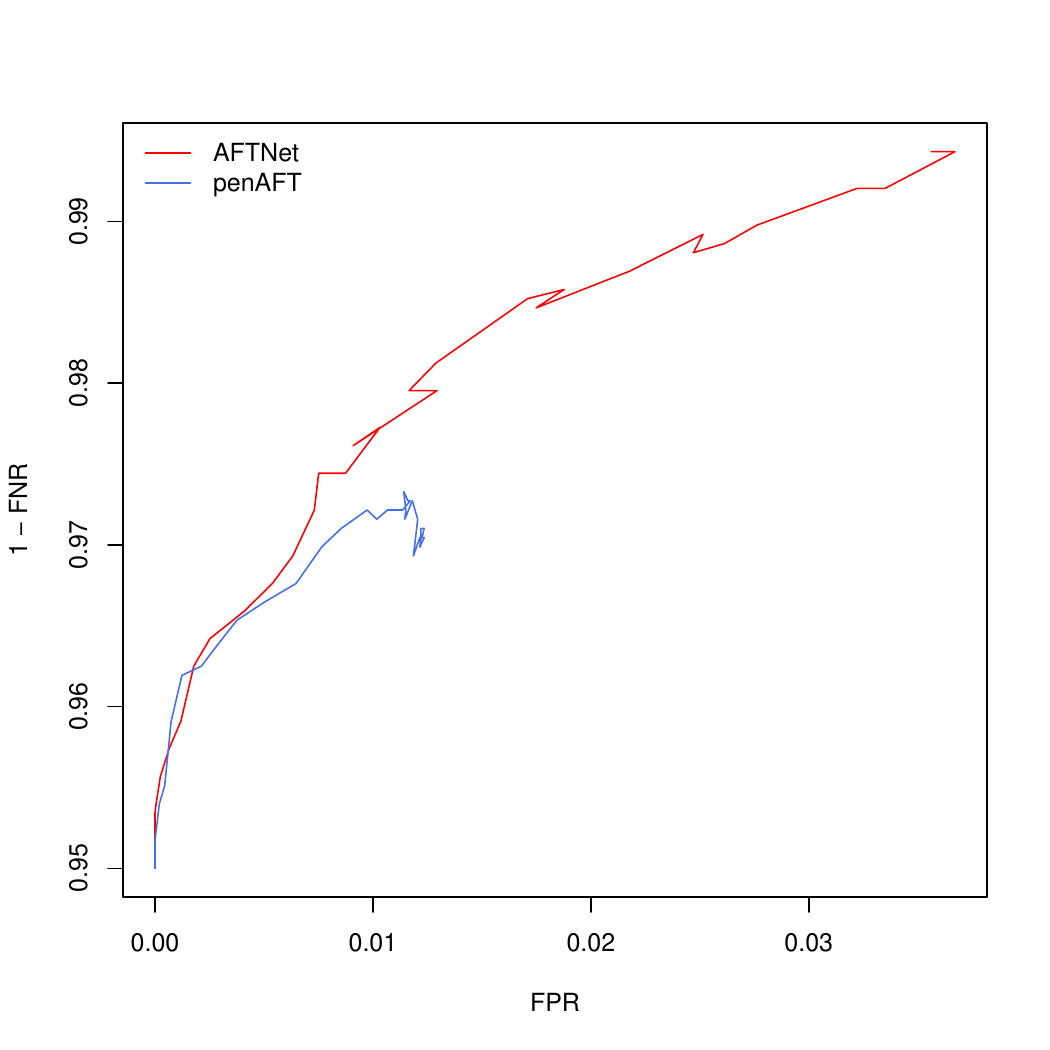}
	\caption{{\bf Overlapping case}. ROC curve for $\sigma=1$ with a weak effect (on the left-side) and a strong effect (on the right-side) .}
	\label{fig6}
\end{figure}

The simulation results for both methods are shown in Table \ref{tab4} and \ref{tab5} for the weak and strong dimensional cases, respectively.  Also, in this case, AFTNet performs better than {\tt penAFT} in terms of EMSE, PMSE, and FNR when a weak or strong dimensional scenario is simulated. Instead,  penAFT tends to detect fewer genes with lower FPR.
\begin{table}[!htbp] \centering 
	\caption{{\bf Overlapping case}. Performance metrics with $n_T=110$, $n_D=55$, $p=220$ and $\alpha=0.5$ (weak effect) averaged over 20 independent replications.} 
	\label{tab4} 
	\begin{tabular}{@{\extracolsep{5pt}} lcccccc} 
		\\[-1.8ex]\hline
		\hline \\[-1.8ex] 
		& \multicolumn{2}{c}{$\sigma=0.5$}   & \multicolumn{2}{c}{$\sigma=1$}  & \multicolumn{2}{c} {$\sigma=1.5$} \\ 
		\hline \\[-1.8ex] 
		& AFTNet & penAFT & AFTNet& penAFT & AFTNet & penAFT\\ 
		\hline \\[-1.8ex] 
		EMSE &  2.686 & 2.798 & 2.809 & 2.930 & 2.884 &	2.982 \\
		PMSE & 0.690 &	0.807 & 0.790 &	0.919 & 0.862 &	0.965 \\
		FNR    &  0.288 & 0.647 & 0.480 & 0.858 & 0.600 & 0.948 \\
		FPR     & 0.539 & 0.152 & 0.364 & 0.069 & 0.297 & 0.019 \\
		NSR    & 0.608 & 0.232 & 0.426 & 0.098 & 0.338 & 0.032 \\
		\hline \\[-1.8ex] 
	\end{tabular} 
\end{table} 
\begin{table}[!htbp] \centering 
	\caption{{\bf Overlapping case}. Performance metrics with $n_T=275$, $n_D=138$, $p=1100$ and $\alpha=0.5$ (strong effect) averaged over 20 independent replications.} 
	\label{tab5} 
	\begin{tabular}{@{\extracolsep{5pt}} lcccccc} 
		\\[-1.8ex]\hline
		\hline \\[-1.8ex] 
		& \multicolumn{2}{c}{$\sigma=0.5$}   & \multicolumn{2}{c}{$\sigma=1$}  & \multicolumn{2}{c} {$\sigma=1.5$} \\ 
		\hline \\[-1.8ex] 
		& AFTNet & penAFT & AFTNet & penAFT & AFTNet & penAFT\\ 
		\hline \\[-1.8ex] 
		EMSE & 2.740 & 2.845 & 2.846 &	2.938 & 2.915 &	2.976 \\
		PMSE & 0.549 &	0.613 & 0.605 &	0.667 & 0.641 &	0.693 \\
		FNR &  0.229 &	0.638 & 0.414 & 0.843 & 0.578 &	0.928 \\
		FPR & 0.436 &	0.065 & 0.294 & 0.034 &  0.226 &	0.019 \\
		NSR &  0.463 & 0.089 & 0.317 &	0.044 & 0.242 &	0.023 \\
		\hline \\[-1.8ex] 
	\end{tabular} 
\end{table} 
A visual summary of  the metrics can be also found in Figures \ref{fig4} and \ref{fig5}.
Moreover, to assess the validity of our method, we show the ROC curve in Figure \ref{fig6} where we can see that AFTnet performs better than penAFT in both weak and strong effects when $\sigma=1$.

Finally, similar to the previous results, the average running time of AFTNet for obtaining a single solution path over 20 independent replications increases compared to penAFT.

\section{Real case studies}\label{section: real}

In this section, we consider two examples of biomarkers discovery related to gene expression data of different types of cancer: breast and kidney. 

%
%
\par 
\begin{itemize}
	\item {\bf Breast cancer data (BC)}: We consider two independent gene expression datasets available from the GEO database (\url{https://www.ncbi.nlm.nih.gov/geo/}). We use GSE2034 as the training set and GSE2990 as the test set and select 13229 genes common to both datasets. The first dataset contains $n_T=286$ records of lymph-node-negative breast cancer patients. The second one contains $n_D=189$ records of invasive breast carcinoma patients. The median survival time (RFS) in the training data set is 86 months with a censoring proportion of 62.59\%, while in the test set, it is 77 months with a censoring proportion equal to  64.17\%.  
	%
	\item {\bf Kidney cancer data (TCGA-KIRC)}: We consider the TCGA-KIRC (Kidney Renal Clear Cell Carcinoma) gene expression dataset available in the GDC Data Portal (\url{https://portal.gdc.cancer.gov}). The data were obtained from the Illumina HiSeq platform, consisting of 518 kidney cancer patients and 20159 genes. We download the pre-processed and normalized data (gene-level, RPKM) from LinkedOmics portal \url{http://linkedomics.org/login.php}. The median survival time (OS) is 1252 days with a censoring proportion of 66.60\%. The data were randomly split into training $T$ and test set $D$ using a re-sampling method, which provides more accurate estimates of predictive accuracy (see, for instance, \cite{simon2011cross} and \cite{iuliano2016cancer}). 
\end{itemize} 

\noindent
As a pre-processing step, we first apply a screening procedure to reduce the dimension of the problem, selecting $p=488$ genes for the data sets of BC and $p=521$ genes for the TCGA-KIRC data set, respectively. The screening process is based on a combination of biological information (BMD) and data-driven knowledge (DAD) (see, for instance, \cite{iuliano2018combining}). In particular, we perform this processing step using the {\tt R} package {\tt COSMONET} \citep{iuliano2021cosmonet}. Subsequently, we conduct inference by running AFTNet and penAFT according to the two real settings investigated (BC and TCGA-KIRC).
To construct the prior network information, we map the screened genes from the KEGG repository (\url{https://www.genome.jp/kegg/pathway.html}) and build a gene adjacency matrix (see \cite{iuliano2021cosmonet} for details). 

During the training phase in Algorithm \ref{alg:1} we initialize $\boldsymbol{\beta}^{(0)}=\boldsymbol{0}$, we fix $\alpha=0.5$ and $M$ the largest eigenvalue of the likelihood Hessian matrix evaluated in $(\boldsymbol{\beta}^{(0)},\hat{\sigma})$. We select the optimal parameter $\lambda_{opt} \in [\lambda_{min},\lambda_{max}]$ by the CV-LP approach illustrated in subsection \ref{section: cross} with $K=5$ folds, $\lambda_{max}= \left\| \nabla \ell \left(\boldsymbol{\beta}^{(0)}, \hat{\sigma} \right) \right\|_{\infty} / \alpha$ and  $\lambda_{min}=0.01 \cdot \lambda_{max}$. 
For the $\lambda$'s grid we consider 50 equispaced points $\zeta_i$ in the interval $[\log10(\lambda_{min}),\log10(\lambda_{max})]$ and take $\lambda_i=10^{\zeta_i}$.

As regards the implementation of the regularized Gehan estimator, we use the {\tt penAFT} package with the following choice $penalty=$``EN'', which stands for elastic-net penalty, with $\alpha=0.5$, $nlambda=50$ different value of regularization parameter and $lambda.ratio.min=0.01$.
In the testing phase, to assess the performance of both methods, we compute both concordance or Harrell’s C-index \citep{harrell1984regression} and the integrated AUC measure (using the {\tt survAUC} package in {\tt R}) on the test set. 

\vspace{0.5cm}
\noindent
{\bf Quantitative results}

The performance measures in the two examples of cancer data are displayed in Table \ref{tab6}. We observe that AFTNet performs better than penAFT in terms of concordance and integrated AUC in the BC data set, while in the TCGA-KIRC data set, AFTNet performs better than penAFT only in terms of AUC.
Interestingly, the running time of AFTNet is slightly higher than penAFT in the BC data set, while it is much smaller in the TCGA-KIRC data set compared to penAFT.

\begin{table}[!htbp] \centering 
	\caption{Harrell’s C-index, integrated AUC, and computing time for AFNet and penAFT methods over the two cancer datasets. The running time includes the time taken for performing 5-fold cross-validation (CV-LP) and model fitting to the complete training dataset with $\alpha=0.5$.} 
	\label{tab6} 
	\begin{tabular}{@{\extracolsep{5pt}} lcccccc} 
		\\[-1.8ex]\hline
		\hline \\[-1.8ex] 
		& \multicolumn{2}{c}{Harrell’s C-index}   & \multicolumn{2}{c}{Integrated AUC}  & \multicolumn{2}{c}{Running time (secs)}   \\ 
		\hline \\[-1.8ex] 
		& AFTNet & penAFT &   AFTNet & penAFT &  AFTNet & penAFT  \\ 
		\hline \\[-1.8ex] 
		BC      & 0.672 & 0.639& 0.396 & 0.325 & 37.388 & 36.278  \\
		TCGA-KIRC  & 0.681 &	0.722 & 0.279 &	0.251 &  49.436 & 283.994\\
		\hline \\[-1.8ex] 
	\end{tabular} 
\end{table} 

\vspace{0.5cm}
\noindent
{\bf Pathway analysis}

To investigate the gene signature, i.e., the set of genes whose corresponding estimated coefficient in $\hat{\boldsymbol{\beta}}$ is non-zero, we perform a pathway analysis based on the information from the KEGG database (\url{https://www.genome.jp/kegg/pathway.html}). In particular, we use the {\tt COSMONET} package (see \cite{iuliano2021cosmonet}) to generate the sub-networks involved in the cancer mechanism and to identify the set of active pathways.
\par 
Before carrying out the KEGG pathway analysis,  we first sort the list of the selected genes of each method (AFTnet and penAFT) in descending order, according to the regression coefficients. Then, we select the top-ranked and bottom-ranked genes into positive (top 20\%) and negative (bottom 20\%) genes. We perform the pathway analysis using not-isolated genes (i.e., connected in the adjacency matrix).
\par 
Specifically, for the BC data set, we obtain that, for both methods, many of the selected genes belong to specific pathways as a group of the {\it KEGG pathways in cancer}, {\it KEGG MAPK/ERBB/P53/chemokine signaling pathway} and {\it KEGG cytokine-cytokine receptor interaction pathway}. However, AFTNet also includes the 
{\it KEGG VEGF signaling pathway}. Indeed, the family of the VEGF proteins has been identified as potential biomarkers of BC and indicators of treatment success and patient survival \citep{brogowska2023vascular}. 
\par 
In the TCGA-KIRC data set, for both methods, the not-isolated genes detected are primarily included in {\it KEGG pathways in cancer}, {\it KEGG chemokine/insulin signaling pathway} and {\it KEGG cycle cell}. Moreover, AFTNet contains {\it KEGG NOTCH signaling pathway}, which plays an essential role in kidney development and disease treatments \citep{barak2012role}.

\section{Conclusions}\label{conclusions}

This article proposes AFTNet, a novel network-constraint survival analysis method considering the Weibull accelerated failure time model. AFTNet addresses the problem of high-dimensionality (i.e., $p>>n$) and the strong correlation among variables exploiting a double penalty that promotes both sparsity and grouping effect. We establish the finite sample consistency and present an efficient iterative computational algorithm based on the proximal gradient descent method.

We show AFTNet's effectiveness in simulations and on two real test cases, comparing its performance with penAFT technique \citep{suder2022scalable}. 

Some interesting future research directions include extending the proposed framework to other parametric AFT models, i.e., log-normal and log-logistic, and using other penalties, particularly non-convex ones such as SCAD or MCP.
 
 \vspace{0.5cm}
 \noindent {\bf{Author contributions}}
 
 \noindent {\it{The authors contributed equally to this work.}} 
 
\section*{Acknowledgement}

I. De Feis and D. De Canditiis acknowledge the INdAM-GNCS Project 2023 ``Metodi computazionali per la modellizzazione e la previsione di malattie neurodegenerative". 

A. Iuliano and C. Angelini acknowledge the INdAM-GNCS Project ``Modelli di shock basati sul processo di conteggio geometrico e applicazioni alla sopravvivenza" (CUP E55F22000270001).

C. Angelini e I. De Feis were partially supported by the project REGINA: ``Rete di Genomica Integrata per Nuove Applicazioni in medicina di precisione" - Ministero della salute nell’ambito del Piano Operativo Salute. Traiettoria 3 ``Medicina rigenerativa, predittiva e personalizzata". Linea di azione 3.1 ``Creazione di un programma di medicina di precisione per la mappatura del genoma umano su scala nazionale". 

A. Iuliano is partially supported by Project Tech4You - Technologies for climate change adaptation and quality of life improvement, n. ECS0000009. This work was funded by the Next Generation EU - Italian NRRP, Mission 4, Component 2, Investment 1.5, call for the creation and strengthening of ``Innovation Ecosystems", building ``Territorial R\&D Leaders" (Directorial Decree n. 2021/3277).

\vspace*{1pc}

\noindent {\bf{Conflict of Interest}}

\noindent {\it{The authors have declared no conflict of interest.}}

\vspace*{1pc}
\noindent {\bf Data Availability Statement}

\noindent {\it The data that support the findings of this study are openly available in Gene Expression Omnibus at https://www.ncbi.nlm.nih.gov/geo/ and in GDC Data Portal - National Cancer Institute at https://portal.gdc.cancer.gov. }

\vspace*{1pc}
\noindent {\bf Orcid }
 
\noindent {Claudia Angelini: https://orcid.org/0000-0001-8350-8464}
 
\noindent {Daniela De Canditiis: https://orcid.org/0000-0002-3022-3411}

\noindent {Italia De Feis: https://orcid.org/0000-0002-3694-8202}

\noindent {Antonella Iuliano: https://orcid.org/0000-0001-8541-8120}

\section*{Appendix {\it(Proofs of technical lemmas)}}

%
\begin{propo} [cfr. Prop 2.5 pag.  24 in \cite{book_martin}]\label{def2}
	({Hoeffding bound}) Assume $X_i$, $i=1, \ldots, n$, sub-gaussian random variables with mean $\mu_i$ and parameters $\sigma_i$, $i=1, \ldots, n$. Then,  for all $t \geq 0,$ we have
	\begin{equation*}
		\mathbb{P}\Bigg[\left|  \sum_{i=1}^{n}(X_i-\mu_i)\right| \geq t\Bigg] \leq 2 \exp\Bigg({-\frac{t^2}{\sum_{i=1}^{n}\sigma_i^2}}\Bigg).
		\label{eq:sHoeffding bound}
	\end{equation*}
	In particular,  for bounded random variables, $X_i \in [a,b]$ such that $\sigma=\frac{b-a}{2}$, we have 
	\begin{equation}
		\mathbb{P}\Bigg[\left|  \sum_{i=1}^{n}(X_i-\mu_i)\right| \geq t\Bigg] \leq 2 \exp\Bigg({-\frac{2t^2}{n(b-a)^2}}\Bigg).
		\label{eq:sHoeffding bound2}
	\end{equation}
\end{propo}

\vspace{0.5cm}

{ \bf Proof of Lemma \ref{lemma1}}

\vspace{0.5cm}

For the integral definition of the mean value theorem generalized to vector-valued functions for some $u \in [0,1]$, we have
\begin{equation*}
	\langle \nabla \ell^{(n)}(\boldsymbol{\theta}^*+\Delta \boldsymbol{\theta})-\nabla\ell^{(n)}(\boldsymbol{\theta}^*),\Delta \boldsymbol{\theta}\rangle=\Delta \boldsymbol{\theta}^T\, \int_{0}^{1}\nabla^2\ell^{(n)}(\boldsymbol{\theta}^* + u\, \Delta \boldsymbol{\theta}){\rm d}u\, \Delta \boldsymbol{\theta},
	\label{eq:lem1.1}
\end{equation*}
for $\lVert \Delta \boldsymbol{\theta} \rVert_1 \leq 2 R$ and where the matrix integral is element-wise. The right-hand side of the equality can be decomposed as $\mathrm{TERM1} + \mathrm{TERM2}$ where
%
\begin{equation}
	\mathrm{TERM1}=\Delta \boldsymbol{\theta}^T \int_{0}^{1}\mathbb{E}_{XY}\left( \nabla^2\ell^{(n)}(\boldsymbol{\theta}^* + u\, \Delta \boldsymbol{\theta})\right){\rm d}u\Delta \boldsymbol{\theta},
	\label{eq:lem1.3}
\end{equation}
\begin{equation}
	\mathrm{TERM2}=\Delta \boldsymbol{\theta}^T \bigg\{\int_{0}^{1}\nabla^2\ell^{(n)}(\boldsymbol{\theta}^*+ u\, \Delta \boldsymbol{\theta})-\mathbb{E}_{XY}\left( \nabla^2\ell^{(n)}(\boldsymbol{\theta}^* + u\, \Delta \boldsymbol{\theta})\right){\rm d}u\bigg\}\Delta \boldsymbol{\theta}.
	\label{eq:lem1.4}
\end{equation}
Let's consider first Eq.\,(\ref{eq:lem1.3}). Note that $\lVert \boldsymbol{\theta}^*-(\boldsymbol{\theta}^* + u \Delta\boldsymbol{\theta} )\rVert_1=u \lVert \Delta \boldsymbol{\theta} \rVert_1 \leq 2R$.  Hence, it holds Asummption \ref{assump3} and we have:
\begin{equation*}
\mathrm{TERM1}	= \Delta \boldsymbol{\theta}^T \int_{0}^{1}\mathbb{E}_{XY}\left( \nabla^2\ell^{(n)}(\boldsymbol{\theta}^* + u\, \Delta \boldsymbol{\theta})\right) {\rm d}u\Delta \boldsymbol{\theta} \geq \gamma \Delta \boldsymbol{\theta}^T \Delta \boldsymbol{\theta}= \gamma \lVert \Delta \boldsymbol{\theta}\rVert_2^2.
	\label{eq:lem1.5}
\end{equation*}
Now consider Eq.\,(\ref{eq:lem1.4}) and define
\begin{equation*}
	G^{\Delta \boldsymbol{\theta} }(u)=\nabla^2\ell^{(n)}(\boldsymbol{\theta}^*+ u\, \Delta \boldsymbol{\theta})-\mathbb{E}_{XY}\left( \nabla^2\ell^{(n)}(\boldsymbol{\theta}^* + u\, \Delta \boldsymbol{\theta}) \right),
	\label{eq:lem1.6}
\end{equation*}
then, Eq.\,(\ref{eq:lem1.4}) became 
\begin{equation*}
	\mathrm{TERM2}=\Delta \boldsymbol{\theta}^T \int_{0}^{1}G^{\Delta \boldsymbol{\theta} }(u){\rm d}u\Delta \boldsymbol{\theta}.
	\label{eq:lem1.7}
\end{equation*}
Therefore, we have by H\"older's inequality
\begin{equation*}
	\begin{aligned}
		\left| \mathrm{TERM2}\right|& =\left|\Delta \boldsymbol{\theta}^T \int_{0}^{1}G^{\Delta \boldsymbol{\theta} }(u){\rm d}u\Delta \boldsymbol{\theta}\right| \leq \lVert \Delta \boldsymbol{\theta}\rVert_1 \, \left\| \int_{0}^{1}G^{\Delta \boldsymbol{\theta} }(u){\rm d}u\Delta \boldsymbol{\theta}\right\|_{\infty}\\
		&=\lVert \Delta \boldsymbol{\theta}\rVert_1 \max_{1 \leq k \leq p+1} \left| \sum_{j=1}^{p+1}\bigg(\int_{0}^{1}G^{\Delta \boldsymbol{\theta} }(u){\rm d}u\bigg)_{kj}\Delta \boldsymbol{\theta}_j\right| \\
		&\leq \lVert \Delta \boldsymbol{\theta}\rVert_1 \max_{1 \leq k \leq p+1}\left\| \bigg(\int_{0}^{1}G^{\Delta \boldsymbol{\theta} }(u){\rm d}u\bigg)_{k \cdot} \right\|_{\infty} \, \lVert \Delta \boldsymbol{\theta}\rVert_1\\
		&=\lVert \Delta \boldsymbol{\theta}\rVert_1^2 \max_{1 \leq k,j \leq p+1}
		\left|\bigg(\int_{0}^{1}G^{\Delta \boldsymbol{\theta} }(u){\rm d}u\bigg)_{k j}\right|\\
		& \leq \lVert \Delta \boldsymbol{\theta}\rVert_1^2 \max_{1 \leq k,j \leq p+1}
		\bigg(\int_{0}^{1} \left|G^{\Delta \boldsymbol{\theta} }(u)\right|{\rm d}u\bigg)_{k j}\\
		& \leq \lVert \Delta \boldsymbol{\theta}\rVert_1^2 \sup_{u \in [0,1]}\max_{1 \leq k,j \leq p+1}\left|G^{\Delta \boldsymbol{\theta}}_{kj}(u)\right|.
	\end{aligned}
	\label{eq:lem1.8}
\end{equation*}
Using an $\varepsilon$-net argument, we define a grid of points for $u$, i.e., $u_m=m/n$, $m=1, \dots,n$. Then,  $\forall u \in [0,1],\,, \exists m: \lvert u-u_m \rvert \leq 1/n$ and
\begin{equation*}
	\begin{aligned}
		\left| \mathrm{TERM2}\right| \leq
		 \underbrace{\lVert \Delta \boldsymbol{\theta}\rVert_1^2 \max_{1 \leq m \leq n}\max_{1 \leq k,j \leq p+1} \left|G^{\Delta \boldsymbol{\theta}}_{kj}(u_m)\right|}_{\mathrm{TERM2}A}+\underbrace{\lVert \Delta \boldsymbol{\theta}\rVert_1^2  \sup_{\lvert u-u_m \rvert \leq 1/n} \max_{1 \leq k,j \leq p+1} \left|G^{\Delta \boldsymbol{\theta}}_{kj}(u)-G^{\Delta \boldsymbol{\theta}}_{kj}(u_m)\right|}_{\mathrm{TERM2}B}
	\end{aligned}
	\label{eq:lem1.9}
\end{equation*}
We consider first $\mathrm{TERM2}B$. Note that $G^{\Delta \boldsymbol{\theta}}_{kj}(u)$ is continuous in $u \in [0,1]$, $\forall k,j$ and therefore locally Lipschitz over $[0,1]$. So, we have
\begin{equation*}
	\left| \mathrm{TERM2}B\right| \leq C \sup_{\lvert u-u_m \rvert \leq 1/n}  \lvert u-u_m \rvert \, \lVert \Delta \boldsymbol{\theta}\rVert_1^2 = \frac{C}{n}\lVert \Delta \boldsymbol{\theta}\rVert_1^2,
	\label{eq:lem1.10}
\end{equation*}
where $C$ is a Lipschitz constant. 
We use the Hoeffding inequality for $\mathrm{TERM2}A$.  Let's consider for all $m,k,j$ fixed
\begin{equation*}
	\begin{aligned}
		\left|G^{\Delta \boldsymbol{\theta}}_{kj}(u_m)\right|&=\left|\Bigg\{\nabla^2\ell^{(n)}(\boldsymbol{\theta}^* + u_m\, \Delta \boldsymbol{\theta})-\mathbb{E}_{XY}\bigg(\nabla^2\ell^{(n)}(\boldsymbol{\theta}^* + u_m\, \Delta \boldsymbol{\theta})\bigg)\Bigg\}_{kj}\right| \\
		&=\left|-\frac{1}{n}\Bigg\{\sum_{i=1}^{n}  \left[ \nabla^2\ell_i(\boldsymbol{\theta}^* + u_m\, \Delta \boldsymbol{\theta})-\mathbb{E}_{XY}\bigg(\nabla^2\ell_i(\boldsymbol{\theta}^* + u_m\, \Delta \boldsymbol{\theta})\bigg)   \right]\Bigg\}_{kj}\right|\\
		&=\left|\frac{1}{n} \sum_{i=1}^{n} \bigg[Z_i- \mathbb{E}(Z_i)\bigg]\right|,
	\end{aligned}
	\label{eq:lem1.11}
\end{equation*}
with $Z_i=\left[\nabla^2\ell_i(\boldsymbol{\theta}^* + u_m\, \Delta \boldsymbol{\theta})\right]_{kj}$ bounded and independent random variables,  implying $Z_i$ sub-gaussian variables. Now, we can use Eq. \ (\ref{eq:sHoeffding bound2})
\begin{equation*}
	\begin{aligned}
		\mathbb{P}\big(\left|G^{\Delta \boldsymbol{\theta}}_{kj}(u_m)\right| \geq t\big)=	\mathbb{P}\left( \left|\frac{1}{n} \sum_{i=1}^{n} \bigg[Z_i- \mathbb{E}(Z_i)\bigg]\right| \geq t\right) \leq 2\exp\bigg(-\frac{n t^2}{c}\bigg),
	\end{aligned}
	\label{eq:lem1.12}
\end{equation*}
where $c$ is the sub-gaussian parameter.
Therefore,  from
\begin{equation*}
	\mathrm{TERM2}A=\lVert \Delta \boldsymbol{\theta}\rVert_1^2 \max_{1 \leq m \leq n}\max_{1 \leq k,j \leq p+1} \left|G^{\Delta \boldsymbol{\theta}}_{kj}(u_m)\right|,
	\label{eq:lem1.13}
\end{equation*}
it follows
\begin{equation*}
	\begin{aligned}
		\mathbb{P}\bigg(\max_{1 \leq m \leq n}\max_{1 \leq k,j \leq p+1} \left|G^{\Delta \boldsymbol{\theta}}_{kj}(u_m)\right| \geq t\bigg)&=\mathbb{P}\bigg(\bigcup_{m=1}^{n} \bigcup_{k.j=1}^{p+1}\left|G^{\Delta \boldsymbol{\theta}}_{kj}(u_m)\right| \geq t\bigg)\\
		&\leq \sum_{m=1}^{n} \sum_{k.j=1}^{p+1}\mathbb{P}\bigg(\left|G^{\Delta \boldsymbol{\theta}}_{kj}(u_m)\right| \geq t\bigg)\\
		&\leq \sum_{m=1}^{n} \sum_{k.j=1}^{p+1} 2 \exp\bigg(-\frac{n t^2}{c}\bigg)\\
		&=2 n(p+1)^2 \exp\bigg(-\frac{n t^2}{c}\bigg).
	\end{aligned}
	\label{eq:lem1.14}
\end{equation*}
We define $\varepsilon=2n(p+1)^2 \exp\bigg(-\frac{n t^2}{c}\bigg)$.  Thus, we get
\begin{equation*}
	\exp\bigg(-\frac{n t^2}{c}\bigg)=\frac{\varepsilon}{2n(p+1)^2}
	\quad  \Rightarrow	\quad	t=\sqrt{\Bigg[\log \frac{2n(p+1)^2}{\varepsilon}\Bigg] \,\frac{c}{n}}.
	\label{eq:lem1.15}
\end{equation*}

Therefore, for all $\varepsilon >0$,
\begin{equation*}
	\begin{aligned}
		\mathbb{P}\Bigg(\mathrm{TERM2}A \leq \sqrt{\Bigg[\log \frac{2n(p+1)^2}{\varepsilon}\Bigg] \,\frac{c}{n}}\Bigg) \geq 1-\varepsilon,
	\end{aligned}
	\label{eq:lem1.17}
\end{equation*}
hence, 
\begin{equation*}
	\begin{aligned}
		\left| \mathrm{TERM2} \right|&\leq \lVert \Delta \boldsymbol{\theta}\rVert_1^2 \, \frac{C}{n}+\lVert \Delta \boldsymbol{\theta}\rVert_1^2 \, \sqrt{\Bigg[\log \frac{2n(p+1)^2}{\varepsilon}\Bigg] \,\frac{c}{n}}\\
		& \leq \lVert \Delta \boldsymbol{\theta}\rVert_1^2  \Bigg\{\frac{C}{n}+\sqrt{ \frac{c \log \left( 2(p+1) / \varepsilon \right)}{n}}+\sqrt{ \frac{c \log \left( n(p+1) \right)}{n}}\Bigg\}\\
	\end{aligned}
	\label{eq:lem1.18}
\end{equation*}
with probability at least $1-\varepsilon$. Finally,  combining all terms,  it follows that it exists $\tau >0$, such that
\begin{equation*}
		\langle \nabla \ell^{(n)}(\boldsymbol{\theta}^*+\Delta \boldsymbol{\theta})-\nabla\ell^{(n)}(\boldsymbol{\theta}^*),\Delta \boldsymbol{\theta}\rangle=\mathrm{TERM1}+\mathrm{TERM2} \geq  \gamma \, \lVert \Delta \boldsymbol{\theta}\rVert_2^2 - \tau \, \lVert \Delta \boldsymbol{\theta}\rVert_1^2\, \sqrt{\frac{\log n(p+1)}{n}},
	\label{eq:lem1.19}
\end{equation*}
with probability at least $1-\varepsilon$.
\begin{flushright}
	$\square$
\end{flushright}

\vspace{0.5cm}

{\bf Proof of Lemma \ref{lemma2}}

To obtain Eq.\,(\ref{lem2})  is sufficient to prove that 
\begin{equation}
	\left\| \nabla\ell^{(n)}(\boldsymbol{\theta}^*)\right\|_{\infty} =O_p\Bigg(\sqrt{\frac{\log \left(  n(p+1) \right) }{n}}\Bigg).
	\label{eq:lem2.1}
\end{equation}
From  Eq.\,(\ref{eq:gradiente}), for all $j=1, \ldots, p+1$ and $t \geq 0$, we have
\begin{equation*}\label{eq:lem2.2}
		\mathbb{P}\Bigg(\left|\frac{\partial \ell^{(n)}(\boldsymbol{\theta}^*)}{\partial \boldsymbol{\theta}_j}\right| \geq t\Bigg)=\mathbb{P}\Bigg(\left|-\frac{1}{n} \sum_{i=1}^{n}\frac{\partial \ell_i(\boldsymbol{\theta}^*)}{\partial \boldsymbol{\theta}_j}\right| \geq t\Bigg)=\mathbb{P}\Bigg(\left|\frac{1}{n} \sum_{i=1}^{n} Z_ i\right|\geq t \Bigg),
\end{equation*}
with $Z_i=\frac{\partial \ell_i(\boldsymbol{\theta}^*)}{\partial \boldsymbol{\theta}_j}$ sub-gaussian random variables with zero mean being $\mathbb{E}_{XY}(\nabla \ell^{(n)}(\boldsymbol{\theta}^*))=0$.
 Using Eq.\ (\ref{eq:sHoeffding bound2}), we have
\begin{equation*}
	\mathbb{P}\Bigg(\left|\frac{\partial \ell^{(n)}(\boldsymbol{\theta}^*)}{\partial \boldsymbol{\theta}_j}\right| \geq t\Bigg) \leq  2 \exp\Bigg(-\frac{n t^2}{c}\Bigg),
	\label{eq:lem2.3}
\end{equation*}
where $c$ is the sub-gaussian parameter. Hence, we get
\begin{equation*}
	\begin{aligned}
		\mathbb{P}\Bigg(\lVert \nabla \ell^{(n)}(\boldsymbol{\theta}^*)\rVert_{\infty}\geq t\Bigg) &= \mathbb{P}\Bigg(\max_{1 \leq j \leq p+1} \left|\frac{\partial \ell^{(n)}(\boldsymbol{\theta}^*)}{\partial \boldsymbol{\theta}_j}\right|\geq t\Bigg)=\mathbb{P}\Bigg(\bigcup_{1 \leq j \leq p+1} \left|\frac{\partial \ell^{(n)}(\boldsymbol{\theta}^*)}{\partial \boldsymbol{\theta}_j}\right|\geq t\Bigg)\\
		& \leq \sum_{j=1}^{p+1}\mathbb{P}\Bigg(\left|\frac{\partial \ell^{(n)}(\boldsymbol{\theta}^*)}{\partial \boldsymbol{\theta}_j}\right|\geq t\Bigg) \leq \sum_{j=1}^{p+1} 2 \exp\Bigg(-\frac{n t^2}{c}\Bigg)=2(p+1) \exp\Bigg(-\frac{n t^2}{c}\Bigg).
	\end{aligned}
	\label{eq:lem2.4}
\end{equation*}
We define $\varepsilon=2(p+1)\exp\bigg(-\frac{n t^2}{c}\bigg)$, thus, we get
\begin{equation*}
		\exp\bigg(-\frac{n t^2}{c}\bigg)=\frac{\varepsilon}{2(p+1)} \quad \Rightarrow \quad t=\sqrt{\frac{c}{n} \log \left( \frac{2(p+1)}{\varepsilon} \right)}.
	\label{eq:lem2.5}
\end{equation*}
This implies that
\begin{equation*}
	\mathbb{P}\Bigg(\lVert \nabla \ell^{(n)}(\boldsymbol{\theta}^*)\rVert_{\infty}\leq \sqrt{\frac{c}{n} \log  \left(  \frac{2(p+1)}{\varepsilon} \right)} \Bigg)  \geq 1-\varepsilon,
	\label{eq:lem2.7}
\end{equation*}
consequently it holds Eq.\,(\ref{eq:lem2.1}).

\begin{flushright}
	$\square$
\end{flushright}



\begin{thebibliography}{10}

\bibitem[Alam {\it et al}.(2022)]{alametal2022}
Alam, T.F., Rahman, M.S., Bari, W. (2022). On estimation for accelerated failure time models with small or rare event survival data. (2022) \textit{BMC Med Res Methodol.}, 22, 169. https://doi.org/10.1186/s12874-022-01638-1
	
	\bibitem[Antoniadis {\it et al}.(2010)]{antoniadisetal2010}
	Antoniadis, A., Fryzlewicz, P., Letu\'{e}, F.  (2010). The Dantzig selector in
Cox's proportional hazards model. \textit{Scandinavian journal of statistics}, 37 (4), 531-552. https://doi.org/10.1111/j.1467-9892.2008.00586.x

	\bibitem[Barak {\it et al}.(2012)]{barak2012role}
	Barak, H., Surendran, K., Boyle, S. C. (2012). The role of Notch signaling in kidney development and disease. \textit{Notch Signaling in Embryology and Cancer}, 99-113.  https://doi.org/10.1007/978-1-4614-0899-4\_8
	
	\bibitem[Barnwal {\it et al}.(2022)]{barnwal2022survival}
	Barnwal, A., Cho, H., Hocking, T. (2022). Survival regression with accelerated failure time model in XGBoost. \textit{Journal of Computational and Graphical Statistics}, 31(4), 1292-1302.
	https://doi.org/10.1080/10618600.2022.2067548
	
	\bibitem[Beck, A. (2017)] {Beck2017} Beck, A. (2017). First order methods in optimization. MOS-SIAM Series on Optimization.
	
	\bibitem[Benner {\it et al}.(2020)]{benneretal2010}	
Benner A., Zucknick M., Hielscher T., Ittrich C., Mansmann U. (2010). High-dimensional Cox models: the choice of penalty as part of the model building process. \textit{Biom J.} , 52(1), 50-69. https://doi.org/10.1002/bimj.200900064. 	
	
	\bibitem[Brogowska {\it et al}.(2023)]{brogowska2023vascular}
	Brogowska, K. K., Zajkowska, M., Mroczko, B. (2023). Vascular Endothelial Growth Factor Ligands and Receptors in Breast Cancer. \textit{Journal of Clinical Medicine}, 12(6), 2412. 
	https://doi.org/10.3390/jcm12062412
	
	\bibitem[Cai {\it et al.}(2009)]{caietal2009}
	Cai, T., Huang, J., Tian, L. (2009) Regularized estimation for the accelerated failure time model. \textit{Biometrics}, 65, 394–404. https://doi.org/10.1111/j.1541-0420.2008.01074.x
	
	\bibitem[Candes and Tao(2007)]{candes2007dantzig}
	Candes E. and Tao T.  (2007). The Dantzig selector: Statistical estimation when p is much larger than n (with discussion). \textit{The Annals of Statistics}, 35, 2313-2404.
	https://doi.org/10.1214/009053606000001523
	
	\bibitem[Chen and Guestrin(2016)]{xgboost}
	Chen, T. and Guestrin, C. (2016). XGBoost: A Scalable Tree Boosting System. \textit{In Proceedings of the 22nd ACM SIGKDD International Conference on Knowledge Discovery and Data Mining}, 785–794. New York, NY, USA: ACM. https://doi.org/10.1145/2939672.2939785
	
	\bibitem[Cheng {\it et al.}(2022)]{chengetal2022}
	Cheng, C., Feng, X., Huang, J., Jiao, Y., Zhang, S. (2022) \textit{l}$_0-$Regularized high-dimensional accelerated failure time model. \textit{Computational Statistics \& Data Analysis}, Volume 170, 107430.
https://doi.org/10.1016/j.csda.2022.107430.

	\bibitem[Cox(1972)]{cox1972regression}
	Cox, D. R. (1972). Regression models and life-tables. \textit{Journal of the Royal Statistical Society: Series B (Methodological)}, 34(2), 187-202.    https://doi.org/10.1111/j.2517-6161.1972.tb00899.x
	
	\bibitem[Dai and Breheny(2019)]{dai2019cross}
	Dai B. and Breheny P. (2019) Cross Validation Approaches for Penalized Cox Regression. \textit{arXiv preprint arXiv}:1905.10432.
	
	\bibitem[Datta {\it et al.}(2007)]{dattaetal2007}
Datta, S., Le-Rademacher, J., Datta, S. (2007). Predicting Patient Survival from Microarray Data by Accelerated Failure Time Modeling Using Partial Least Squares and LASSO. \textit{Biometrics}, 63(1), 259–271.  https://doi.org/10.1111/j.1541-0420.2006.00660.x 

	\bibitem[Du {\it et al}.(2010)]{duetal2010}	
Du, P., Ma, S., Liang, H. (2010).  Penalized Variable Selection Procedure for Cox Models with Semiparametric Relative Risk.   \textit{The Annals of Statistics}, 38(4), 2092–2117.  https://doi.org/10.1214/09-AOS780
	
	\bibitem[Engler and Li(2009)]{englerli2009}	
Engler, D. and Li, Y. (2009) Survival analysis with high-dimensional covariates: an application in microarray studies. \textit{Stat. Appl. Genet. Mol. Biol.}, 8(1), 1–22, (Article 14).  https://doi.org/0.2202/1544-6115.1423 

	\bibitem[Fan and Li(2001)]{fan2001variable}
	Fan, J. and Li, R. (2001). Variable selection via nonconcave penalized likelihood and its oracle properties. \textit{Journal of the American statistical Association}, 96(456), 1348-1360.
	https://doi.org/10.1198/016214501753382273
	
	\bibitem[Fan and Li(2002)]{fan2002variable}
	Fan, J. and Li, R. (2002). Variable selection for Cox's proportional hazards model and frailty model. \textit{The Annals of Statistics}, 30(1), 74-99.	https://doi.org/10.1214/aos/1015362185
	
	\bibitem[Firth(1993)]{firth1993}
	Firth D. Bias reduction of maximum likelihood estimate. (1993) \textit{Biometrika}, 80(1), 27–38. https://doi.org/10.1093/biomet/80.1.27
	
	\bibitem[Frank and Friedman (1993)]{frankfriedman1993}
	Frank, I.E. and Friedman, J.H. (1993). A Statistical View of Some Chemometrics Regression Tools, \textit{Technometrics}, 35, 109–148. https://doi.org/10.1080/00401706.1993.10485033
	
	\bibitem[Friedman and Popescu (2004)]{friedmanpopescu2004}
	Friedman, J. H. and Popescu, B. E. (2004). Gradient directed regularization for linear regression and classification.
\textit{Technical Report}, Department of Statistics, Stanford University, Stanford, California. 
		
	\bibitem[Gong et al.(2014)]{gongetal2014}
Gong, H.; Wu, T.T.; Clarke, E.M. (2014). Pathway-gene identification for pancreatic cancer survival via doubly regularized Cox regression. {\it BMC Syst. Biol.}, 8, 1–9.  https://doi.org/10.1186/1752-0509-8-S1-S3 
		
	\bibitem[Gui and Li(2005)]{gui2005penalized}
	Gui, J. and Li, H. (2005). Penalized Cox regression analysis in the high-dimensional and low-sample size settings, with applications to microarray gene expression data.  \textit{Bioinformatics}, 21(13), 3001-3008.
	https://doi.org/10.1093/bioinformatics/bti422
	
	\bibitem[Harrell {\it et al}.(1984)]{harrell1984regression}
	Harrell Jr, F. E., Lee, K. L., Califf, R. M., Pryor, D. B., Rosati, R. A. (1984). Regression modelling strategies for improved prognostic prediction. \textit{Statistics in Medicine}, 3(2), 143-152.
	 https://doi.org/10.1002/sim.4780030207

\bibitem[Huang and Harrington (2005)]{huangharrington2005}
Huang, J. and Harrington, D. (2005). Iterative Partial Least Squares with Right-Censored Data Analysis: A Comparison to Other Dimension Reduction Techniques. \textit{Biometrics}, 61(1), 17–24. https://doi.org/10.1111/j.0006-341X.2005.040304.x 

\bibitem[Huang {\it et al}.(2006)]{huangetal2006}
Huang J., Ma S., Xie H. (2006). Regularized estimation in the accelerated failure time model with high-dimensional covariates. \textit{Biometrics}, 62(3):813-20. https://doi.org/10.1111/j.1541-0420.2006.00562.x. 
	 
	 \bibitem[Huang and Ma (2010)]{huangma2010}
	 Huang, J. and Ma, S. (2010). Variable selection in the accelerated failure time model via the bridge method. \textit{Lifetime Data Anal.}, 16, 176–195. https://doi.org/10.1111/10.1007/s10985-009-9144-2

	 	\bibitem[Huang {\it et al}.(2014)]{huangetal2014}
Huang, J., Liu, L., Liu, Y., Zhao, X. (2014).  Group selection in the Cox model with diverging number of covariates. . \textit{Statistica Sinica}, 24(4), 1787–1810. https://doi.org/10.5705/ss.2013.061

	 \bibitem[Huang {\it et al}.(2016)]{huangetal2016}
	 	Huang, J., Breheny, P., Lee, S., Ma, S., Zhang, C.-H. (2016). The Mnet method for variable selection.  \textit{Statistica Sinica}, 26(3), 903–923. https://doi.org/10.5705/ss.202014.0011 

 \bibitem[Huang {\it et al}.(2018)]{huangetal2018}
Huang, J., Jiao, Y., Liu, Y., Lu, X. (2018) A constructive approach to \textit{l}$_0$ penalized regression. \textit{J. Mach. Learn. Res.} , 19 (1), 403-439

 \bibitem[Hutton and Monaghan(2002)]{huttonmonaghan2002}
Hutton, J.L. and Monaghan, P.F. (2002) Choice of Parametric Accelerated Life and Proportional Hazards Models for Survival Data: Asymptotic Results. \textit{Lifetime Data Anal}, 8, 375–393 (2002). https://doi.org/10.1023/A:1020570922072	
	
	\bibitem[Iuliano {\it et al}.(2016)]{iuliano2016cancer}
	Iuliano, A., Occhipinti, A., Angelini, C., De Feis, I., Lió, P. (2016). Cancer markers selection using network-based cox regression: a methodological and computational practice. \textit{Frontiers in Physiology}, 7, 208.
	https://doi.org/10.3389/fphys.2016.00208
	
	\bibitem[Iuliano {\it et al}.(2018)]{iuliano2018combining}
	Iuliano, A., Occhipinti, A., Angelini, C., De Feis, I., Liò, P. (2018). Combining pathway identification and breast cancer survival prediction via screening-network methods. \textit{Frontiers in Genetics}, 9, 206.
	https://doi.org/10.3389/fgene.2018.00206	
	
	\bibitem[Iuliano {\it et al}.(2021)]{iuliano2021cosmonet}
	Iuliano, A., Occhipinti, A., Angelini, C., De Feis, I., Li\'{o}, P. (2021). Cosmonet: An R package for survival analysis using screening-network methods. \textit{Mathematics}, 9(24), 3262.
	 https://doi.org/10.3390/math9243262

	\bibitem[Jiang and Liang(2018)]{jiangliang2018}
	Jiang, H.K. and Liang, Y. (2018). The L1/2 regularization network Cox model for analysis of genomic data. \textit{Comput. Biol. Med.}, 100, 203–208. https://doi.org/10.1016/j.compbiomed.2018.07.009 

   \bibitem[Khan and Shaw (2013)]{khanshaw2013}
    Khan, M.H.R. and Shaw, J.E.H. (2013) Variable selection with the modified Buckley- James method and the dantzig selector for high-dimensional survival data. \textit{In: 59th ISI World Statistics Congress Proceedings}, Hong Kong, pp. 4239–4244, 25–30 Aug 2013c

	\bibitem[Kim {\it et al}.(2012)]{kimetal2012}
		Kim J, Sohn I, Jung SH, Kim S, Park C. (2012). Analysis of survival data with group lasso. \textit{Commun. Stat. Simul. Comput.}, 41(9), 1593–1605. https://doi.org/10.1080/03610918.2011.611311
	
	\bibitem[Li and Li (2010)]{lili2010}	
		Li, C. and Li, H. (2010). Variable selection and regression analysis for covariates with graphical structure. \textit{Ann. Appl. Stat.}, 4, 1498–1516. https://doi.org/10.1214/10-AOAS332
			
    \bibitem[Li {\it et al}.(2021)]{Li et al2021} 
    Li, R., Tanigawa Y.,  Justesen Y.M., Taylor, J., Hastie, T.,  Tibshirani, R., Rivas, M.A. (2021) Survival analysis on rare events using group-regularized multi-response Cox regression. \textit{Bioinformatics}, 37(23), 4437–4443. doi: 10.1093/bioinformatics/btab095
	
	\bibitem[Liu(2018)]{liu2018using}
	Liu, E. (2018). Using weibull accelerated failure time regression model to predict survival time and life expectancy. \textit{BioRxiv}, 362186. https://doi.org/10.1101/362186
	
    \bibitem[Loh and Wainwright(2015)]{Loh&Wain2015}	
    Loh, P. L. and Wainwright, M. J. (2015). Regularized M-estimators with Nonconvexity: Statistical and Algorithmic Theory for Local Optima.\textit{Journal of Machine Learning Research}, 16 (19), 559 - 616. 	
		
	\bibitem[Parikh and Boyd(2014)]{parikh2014proximal}
	Parikh, N. and Boyd, S. (2014). Proximal algorithms. \textit{Foundations and Trends{\textregistered} in Optimization}, 1(3), 127-239.
	
	\bibitem[Park and Do(2018)]{park2018penalized}
	Park, E. and Do Ha, I. (2018). Penalized variable selection for accelerated failure time models. \textit{Communications for Statistical Applications and Methods}, 25(6), 591-604.
	https://doi.org/10.29220/CSAM.2018.25.6.591
	
	\bibitem[Reeder {\it et al}.(2023)]{InspirationTheoretical}  
	Reeder, H.T., Lu, J., Haneuse, S. (2023) Penalized estimation of frailty-based illness–death models for semi-competing risks. \textit{Biometrics}, 79, 1657–1669. https://doi.org/10.1111/biom.13761 
	
	\bibitem[Ren {\it et al}.(2019)]{renetal2019}
	Ren J, Du Y, Li S, Ma S, Jiang Y, Wu C. (2019) Robust network-based regularization and variable selection for high-dimensional genomic data in cancer prognosis. \textit{Genet Epidemiol.}, 43(3):276-291. https://doi.org/10.1002/gepi.22194. 
	
     \bibitem[Simon {\it et al}. (2011)]{simon2011cross}
     Simon R.M., Subramanian J., Li M.C., Menezes S.( 2011) Using cross-validation to evaluate predictive accuracy of survival risk classifiers based on high-dimensional data. {\it Briefings in bioinformatics}, 12(3), 203-214.  https://doi.org/10.1093/bib/bbr001
     
    \bibitem[Sha {\it et al}.(2006)]{shaetal2006}
    Sha, N., Tadesse, M.G., Vannucci, M. (2006) Bayesian variable selection for the analysis of microarray data with censored outcome. \textit{Bioinformatics}, 22(18), 2262–2268.  https://doi.org/10.1093/bioinformatics/btl362

	\bibitem[Suder and Molstad(2022)]{suder2022scalable}
	Suder, P. M. and Molstad, A. J. (2022). Scalable algorithms for semiparametric accelerated failure time models in high-dimensions.  \textit{Statistics in Medicine}, 41(6), 933-949.     https://doi.org/10.1002/sim.9264
	
	\bibitem[Sun {\it et al}.(2014)]{sun2014network}
	Sun, H., Lin, W., Feng, R., Li, H. (2014). Network-regularized high-dimensional Cox regression for analysis of genomic data. \textit{Statistica Sinica}, 24(3), 1433.	https://doi.org/10.5705/ss.2012.317
	
	\bibitem[Tibshirani(1996)]{tibshirani1996regression}
	Tibshirani, R. (1996). Regression shrinkage and selection via the lasso.  \textit{Journal of the Royal Statistical Society. Series B (Methodological)}, 58(1), 267-288.	 https://doi.org/10.1111/j.2517-6161.1996.tb02080.x
	
	\bibitem[Tibshirani(1997)]{tibshirani1997lasso}
	Tibshirani, R. (1997). The lasso method for variable selection in the Cox model. \textit{Statistics in medicine}, 16(4), 385-395.	https://doi.org/10.1002/(sici)1097-0258(19970228)16:4<385::aid-sim380>3.0.co;2-3
	
	\bibitem[Verissimo {\it et al}.(2016)]{verissimo2016degreecox}
	Veríssimo, A., Oliveira, A. L., Sagot, M. F., Vinga, S. (2016). DegreeCox - a network-based regularization method for survival analysis. \textit{BMC bioinformatics}, 17, 109-121. 	https://doi.org/10.1186/s12859-016-1310-4
		
	\bibitem[Zhang and Lu (2007)]{zhanglu2007}
	Zhang, H. H. and Lu, W. (2007). Adaptive Lasso for Cox’s Proportional Hazards Model. \textit{Biometrika}, 94(3), 691–703. https://doi.org/10.1093/biomet/asm037
	
	\bibitem[Zhang(2010)]{zhang2010}
	Zhang, C. (2010). Nearly unbiased variable selection under minimax concave penalty. \textit{The Annals of Statistics}, 38(2), 894–942. https://doi.org/10.1214/09‐AOS729.
	
	\bibitem[Zhang {\it et al}.(2013)]{zhang2013network}
	Zhang, W., Ota, T., Shridhar, V., Chien, J., Wu, B., Kuang, R. (2013). Network-based survival analysis reveals subnetwork signatures for predicting outcomes of ovarian cancer treatment.  \textit{PLoS Computational Biology}, 9(3), e1002975.	https://doi.org/10.1371/journal.pcbi.1002975
	
	\bibitem[Zou and Hastie(2005)]{zou2005regularization}
	Zou, H. and Hastie, T. (2005). Regularization and variable selection via the elastic net. \textit{Journal of the Royal Statistical Society: series B (Statistical Methodology)}, 67(2), 301-320.	 https://doi.org/10.1111/j.1467-9868.2005.00503.x
	
	\bibitem[Zou(2006)]{zou2006adaptive}
	Zou, H. (2006). The adaptive lasso and its oracle properties. \textit{Journal of the American Statistical Association} 101(476), 1418-1429. 	https://doi.org/10.1198/016214506000000735
	
	\bibitem[Wainwright(2019) ]{book_martin}  
	Wainwright, M.J. (2019) High-Dimensional Statistics: A Non-Asymptotic Viewpoint. Cambridge Series in Statistical and Probabilistic Mathematics Cambridge University Press.  
	
	\bibitem[Wang {\it et al.}(2008) ]{wangetal2008}  
    Wang, S., Nan, B., Zhu, J., Beer, D. G. (2008). Doubly Penalized Buckley-James Method for Survival Data with High-Dimensional Covariates. {\it Biometrics}, 64(1), 132–140. https://doi.org/10.1111/j.1541-0420.2007.00877.x 

	\bibitem[Wang {\it et al.}(2009) ]{wangetal2009}  
     WANG, S., NAN, B., ZHOU, N., ZHU, J. (2009). Hierarchically penalized Cox regression with grouped variables. \textit{Biometrika}, 96(2), 307–322.  https://doi.org/10.1093/biomet/asp016
	
	\bibitem[Wu and Wang(2013) ]{wuwang2013}  
	Wu, T.T. and  Wang, S. (2013) Doubly Regularized Cox Regression for High-dimensional Survival Data with Group Structures. \textit{Stat. Its Interface}, 6, 175–186.  https://doi.org/10.4310/SII.2013.v6.n2.a2
	
\end{thebibliography}
\end{document}